%% file: main.tex
\documentclass[lettersize,journal]{IEEEtran}
\usepackage{amsmath,amsfonts}
\usepackage{algorithmic}
\usepackage{array}
\usepackage[caption=false,font=normalsize,labelfont=sf,textfont=sf]{subfig}
\usepackage{textcomp}
\usepackage{stfloats}
\usepackage{url}
\usepackage{verbatim}
\usepackage{graphicx}
\usepackage{amssymb}
\usepackage{amsmath}
\usepackage{multirow}
\usepackage{multicol}
\usepackage{booktabs}
\usepackage{pifont}
\usepackage{color}
\usepackage{tcolorbox}
\usepackage{enumitem}
\usepackage{xcolor}
\usepackage[table]{xcolor}

\newcommand{\rev}[1]{\textcolor{black}{#1}}
\newcommand{\revv}[1]{\textcolor{black}{#1}}

\usepackage{newfloat}
\usepackage{listings}

\def\BibTeX{{\rm B\kern-.05em{\sc i\kern-.025em b}\kern-.08em
    T\kern-.1667em\lower.7ex\hbox{E}\kern-.125emX}}
\usepackage{balance}
\usepackage{cite}
\usepackage[colorlinks=true, linkcolor=blue, citecolor=blue, urlcolor=blue]{hyperref}

\renewcommand{\thetable}{\Roman{table}}

\begin{document}
\title{BARE: Towards Bias-Aware and Reasoning-Enhanced \\One-Tower Visual Grounding}
\author{Hongbing Li, Linhui Xiao, Zihan Zhao, Qi Shen, Yixiang Huang, Bo Xiao, 
\\and Zhanyu Ma, \textit{Senior Member, IEEE}
\thanks{
    This research was supported by the National Natural Science Fund of China (Grant Nos.       62076031 and  62076036) and the Major Key Project of PCL under Grant PCL2025A14. (Corresponding
    author: Linhui Xiao and Bo Xiao.)

    Hongbing Li, Zihan Zhao, Qi Shen, Yixiang Huang, Bo Xiao, and Zhanyu Ma are with the School of Artificial Intelligence, Beijing University of Posts and Telecommunications (BUPT), Beijing 100876, China. (e-mail: hbl@bupt.edu.cn, zhaozihan1055@bupt.edu.cn, shenqi@bupt.edu.cn, huangyixiang@bupt.edu.cn, xiaobo@bupt.edu.cn, mazhanyu@bupt.edu.cn)

    Linhui Xiao is with the Pengcheng Laboratory, Shenzhen 518066, China, and also with Institute of Automation, Chinese Academy of Sciences (CASIA), Beijing 100190, China. (e-mail: xiaolinhui16@mails.ucas.ac.cn).
    }
}

\markboth{IEEE TRANSACTIONS ON CIRCUITS AND SYSTEMS FOR VIDEO TECHNOLOGY, DECEMBER 2025}
{BARE: Towards Bias-Aware and Reasoning-Enhanced \\One-Tower Visual Grounding}

\maketitle

\begin{abstract}
\input{Body/0-abstract.tex}

\end{abstract}


\section{Introduction}
\input{Body/1-intro.tex}

\section{Related Work}
\input{Body/2-RelatedWork.tex}

\section{Method}
\input{Body/3-method.tex}

\section{Experiments}

\input{Body/4-Experiments.tex}

\section{Future Work}
\input{Body/5-futurework.tex}

\section{Conclusion}

\input{Body/6-conclusion.tex}

\bibliographystyle{IEEEtran}
\bibliography{references}

\vfill

\end{document}

%% file: Body/0-abstract.tex
Visual Grounding (VG), which aims to locate a specific region referred to by  expressions, is a fundamental yet challenging task in the multimodal understanding fields. 
While recent grounding transfer works have advanced the field through one-tower architectures, they still suffer from two primary limitations: 
\textit{(1)} over-entangled multimodal representations that exacerbate deceptive modality biases, and  
\textit{(2)} insufficient semantic reasoning  that hinders the comprehension of referential cues. 
In this paper, we propose BARE, a bias-aware and reasoning-enhanced framework for one-tower visual grounding. 
BARE introduces a mechanism that preserves modality-specific features and constructs referential semantics through three novel modules: \textit{(i)} language salience modulator, \textit{(ii)} visual bias correction and \textit{(iii)} referential relationship enhancement, which jointly mitigate multimodal distractions and enhance referential comprehension.
Extensive experimental results on five benchmarks demonstrate that BARE not only achieves state-of-the-art performance but also delivers superior computational efficiency compared to existing approaches. 
The code is publicly accessible at \url{https://github.com/Marloweeee/BARE}.

%% file: Body/1-intro.tex

Visual Grounding (VG) \cite{Refcoco,vg_survey}, also known as Referring Expression Comprehension (REC), aims to identify the target regions within an image through a referring expression. 
This task requires a comprehensive perception of the visual context and effective reasoning over fine-grained cross-modal relationships, making it a fundamental yet challenging problem in the field of multimodal understanding. 
Unlike conventional object detection methods that require predefined closed-set categories (\textit{e.g.}, 80 classes in COCO \cite{COCO}), VG is not limited to such fixed \rev{labels}. 
Instead, it flexibly identifies regions based on its semantic understanding of the query expression.
Therefore, serving as a bridge between natural language and the physical world, VG holds potential for a range of applications, including visual dialogue \cite{vqa,generalVQA,VD} and autonomous driving \cite{autonomous_driving}.

\begin{figure}[t]\centering
	\includegraphics[width=8.5cm]{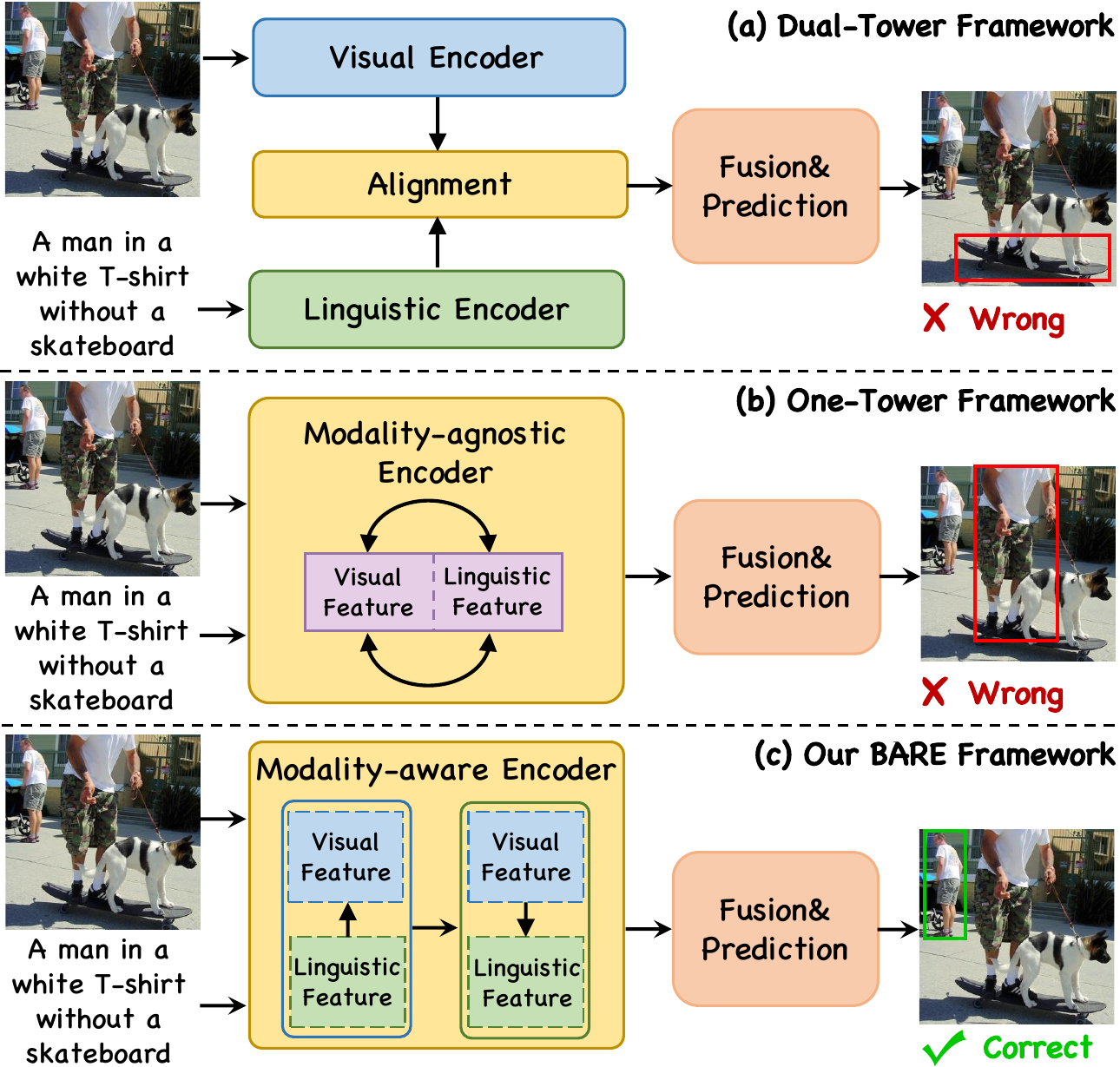}
	\caption{ Comparison of visual grounding frameworks. 
    {(a) Dual-tower architectures employ separate encoders for each modality but often suffer from grounding errors due to the modality gap.}
    (b) One-tower architectures employ a shared encoder, which may lead to feature entanglement and limit grounding performance. 
    {(c) Our BARE leverages decoupled and fully interactive representations for accurate grounding.}
        }
    \label{FIG_1}
    \vspace{-8pt}
\end{figure}


{Building upon powerful pretrained models such as ViT \cite{ViT} and BEiT \cite{BEiT}, most \rev{state-of-the-art (SOTA)} VG methods \cite{vg_survey} adopt the pretraining and fine-tuning paradigm. }
As shown in Fig.\ref{FIG_1}, they first leverage pretrained models to extract multimodal features, followed by fusion and prediction. 
In terms of the architecture, these methods can be categorized into dual-tower and one-tower structures. 
Dual-tower structures \cite{TransVG} leverage pretrained visual models (\textit{e.g.}, ResNet \cite{ResNet}, Swin Transformer \cite{SwinTransformer}) and language models (\textit{e.g.}, BERT \cite{BERT}, BART \cite{bart}) to separately encode visual and linguistic modalities, additionally incorporating intricate mechanisms such as query injection \cite{QRNet}, iterative reasoning \cite{VLTVG}, combination of prompts and adapters \cite{TransVG++}, weight generation \cite{VG-LAW}, cross-modal bridge \cite{HiVG}, etc.
As shown in Fig. \ref{FIG_1}(a), these approaches may result in grounding errors with complex referents, revealing that merely reinforcing representation alignment is insufficient to bridge the modality gap \cite{MindTheGap}.


%
%
%

In contrast, the one-tower \cite{SimVG} framework employs a modality-shared encoder (\textit{e.g.} BEiT \cite{BEiT}) to extract multimodal features, mitigating the feature heterogeneity caused by separated encoders.  
However, the one-tower framework still faces two major challenges.

\begin{figure}[t]\centering
	\includegraphics[width=8.8cm]{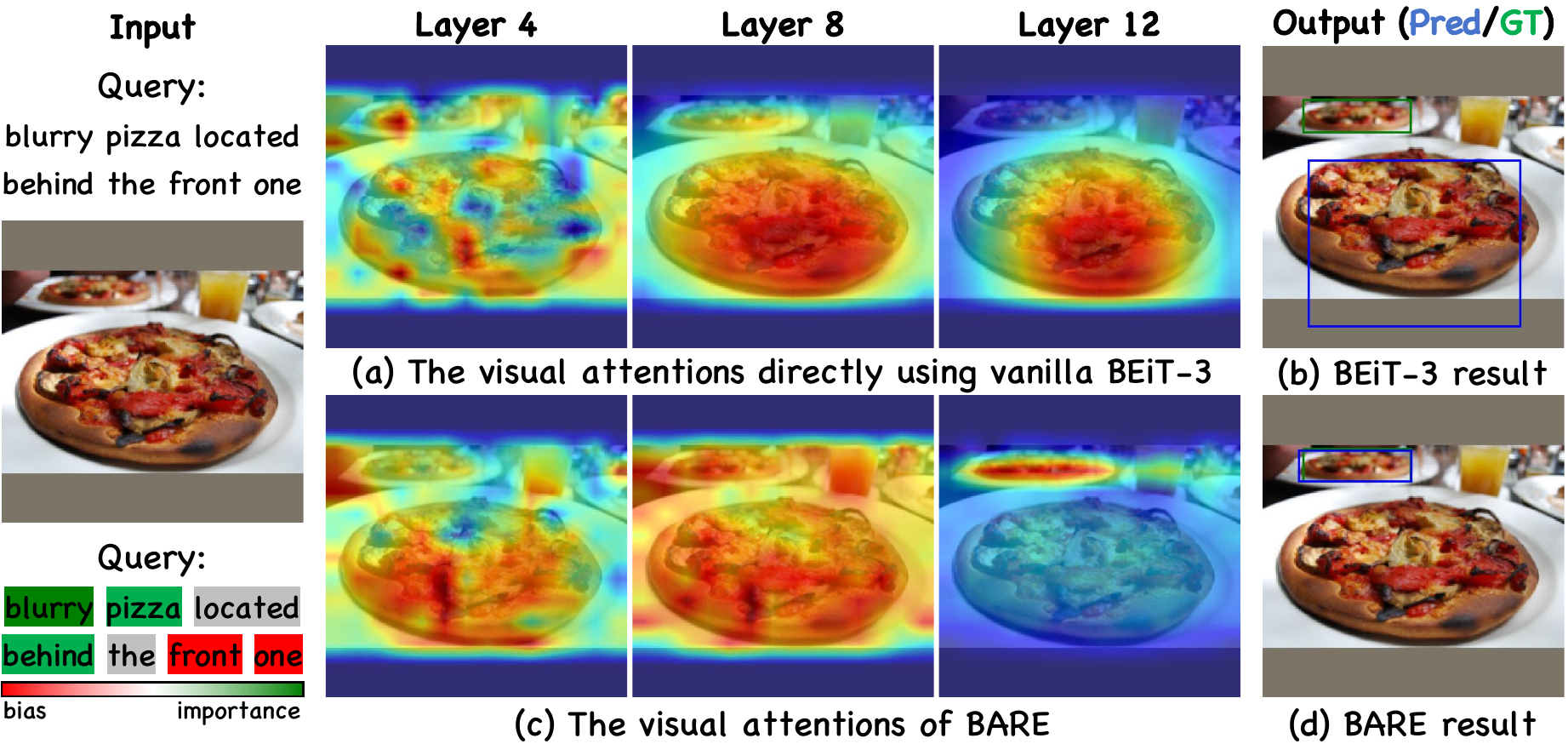}

	\caption{Visualization of layer-wise attention maps and grounding results from BEiT-3 and our proposed method. BARE effectively filters out deceptive visual shortcuts and progressively refines its focus toward the intended referent, leading to more precise and robust grounding under challenging conditions.}    
    \label{FIG_2}
    \vspace{-8pt}
\end{figure}

\textit{\textbf{Firstly}}, the use of a modality-shared encoder, typically implemented via self-attention, introduces an implicit modality bias. As self-attention is inherently modality-agnostic, it tends to homogenize visual and linguistic features within a shared representation space \cite{challenge1}. This excessive blending suppresses modality-specific cues and blurs the distinct contributions of each modality, leading to over-entangled representations. Consequently, the model often struggles to determine whether a prediction is driven by visual evidence or linguistic guidance, degrading its ability to reliably localize the referent, particularly in visually ambiguous scenarios.
\textit{\textbf{Secondly}}, most methods facilitate interaction by simply concatenating multimodal features followed by self-attention. While computationally straightforward, this approach lacks the ability to model structured semantics and thus fails to support compositional reasoning over referential expressions \cite{challenge2, ModalityCompetition}.
As illustrated in Fig. 1(b) and Fig. 2(a), the vanilla model often struggles to interpret complex linguistic constructs, such as attribute dependencies (\rev{\textit{e.g., ``without a skateboard"}}) and abstract concepts ({\textit{e.g., ``blurry pizza"}}). This reasoning deficiency leads to inconsistent visual attention and undermines the model's ability to handle subtle or context-sensitive queries.

Recent efforts attempt to address the above issues through fusion decoupling \cite{SimVG} or hierarchical alignment \cite{YORO}. 
{For instance, while SimVG \cite{SimVG} alleviates feature entanglement, the coupled representations in the encoding stage still limit overall performance.}  
Similarly, YORO \cite{YORO} enables multi-level interaction but often obscures referential semantics, especially under ambiguous scenarios or complex queries.  
\rev{These limitations highlight the necessity of addressing visual bias and reasoning deficiency, preserving the identity of each modality while enabling structured interpretation of referential cues. To this end, we propose a \textbf{B}ias-\textbf{A}ware and \textbf{R}easoning-\textbf{E}nhanced (\textbf{BARE}) framework for one-tower  visual grounding.} 
\textit{\textbf{Firstly}}, we propose Visual Bias Correction (VBC) to mitigate dominant object bias in visual features. 
VBC utilizes learnable semantic prototypes that engage with contextual visual features and selectively incorporate textual cues, thereby recalibrating attention towards target-relevant regions. 
Such targeted filtering accentuates the salience of essential visual attributes while suppressing distracting priors, thereby facilitating more robust and precise grounding in cluttered or biased scenarios. 
\textit{\textbf{Secondly}}, to strengthen reasoning capability, we design the Language Salience Modulator (LSM) and Referential Relationship Enhancement (R$^2$E). 
Specifically, the LSM module performs linguistic salience modulation via a dual-gating mechanism, which adaptively suppresses biased frequency-based shortcuts while accentuating the salience of task-critical referential cues. 
Complementary to this, the R$^2$E module is integrated after the encoder to explicitly model fine-grained semantic dependencies between refined linguistic signals and visual entities, thereby facilitating structured referential reasoning in complex scenarios.
{As shown in Fig. 2(c), BARE first decouples modality-agnostic features and then effectively captures complex referential dependencies, substantially improving grounding accuracy under challenging linguistic conditions.}  
To adapt the pretrained model for grounding tasks, one straightforward approach is full fine-tuning.  However, such practice risks catastrophic forgetting, compromising the retention of general-purpose knowledge.  
Consequently, we leverage Low-Rank Adaptation (LoRA) \cite{LoRA}, which enables efficient and targeted adaptation by tuning a small subset of parameters (only 1.7\%).

The main contributions can be summarized as three-fold:

\begin{itemize}
\item { We propose {BARE}, a bias-aware and reasoning-enhanced one-tower framework for visual grounding, which explicitly mitigates over-entangled features and semantic ambiguity by preserving modality-specific cues and promoting structured cross-modal interaction.} 
\item { {BARE} introduces language salience modulator, visual bias correction and referential relationship enhancement modules that jointly suppress linguistic shortcuts and visual distractions while reinforcing referential dependencies, thereby enabling precise grounding of targets and their contextual relationships in complex queries.}  
\item Extensive experiments on five widely used benchmarks validate the effectiveness of {BARE}. Our model consistently outperforms SOTA methods under the same settings and shows superior computational efficiency.  
\end{itemize}

%% file: Body/2-RelatedWork.tex
\subsection{Visual Grounding}
\rev{Visual grounding studies primarily fall into three categories.}  
\rev{The first involves \revv{typically} applying pretrained close-set detection and language models. As the most traditional approach,} it includes both two-stage and one-stage methods. However, they heavily rely on object detectors, increasingly becoming a performance bottleneck. 
With the advancement of Vision Transformers (ViT) \cite{ViT}, a growing number of studies \cite{TransVG,QRNet,VLTVG,SegVG} have adopted Transformer-based architectures to further enhance grounding performance. 
However, most of these approaches utilize separate visual and language encoders, neglecting the alignment between them. 
Although some works, such as QRNet \cite{QRNet} and M2IST \cite{M2IST}, have introduced text-guided enhancements, they still fall short in establishing structured cross-modal interaction. 

The second category builds upon pretrained dual-tower (\textit{e.g.}, CLIP \cite{CLIP}) or one-tower (\textit{e.g.},  BEiT \cite{BEiT}) architectures, and adapts them to downstream grounding tasks via task-specific fine-tuning. 
Dual-tower approaches, including DUET \cite{VisualGroundingWithDualKnowledgeDistillation} and HiVG \cite{HiVG}, leverage knowledge distillation alongside multi-layer features to facilitate robust cross-modal semantic alignment. However, the use of separate encoders still poses challenges in bridging the inherent modality gap \cite{MindTheGap}. 
In contrast, while one-tower methods \cite{YORO,SimVG} alleviate this issue by enabling early fusion, they often yield highly coupled representations due to the modality-agnostic interactions \cite{challenge1}. 

More recently, with the advancement of large-scale pretraining, a third category has emerged that leverages mixed-dataset fine-tuning on vision-language models. This includes methods based on open-set detection pretraining (\textit{e.g.}, Grounding DINO \cite{Grounding-DINO}) and multi-task mixup-supervised pretraining (\textit{e.g.}, UniTAB \cite{UniTAB} and Grounding GPT \cite{groundinggpt}).
However, these methods rely heavily on large-scale annotations and computational resources, leading to high training costs. 
{In contrast, our method adopts a training-efficient architecture and leverages LoRA-based optimization to significantly improve the usability and effectiveness of a one-tower model.} 
Moreover, our method introduces a novel mechanism for modality debiasing and reasoning, which simultaneously suppresses modality distractions and enhances referential understanding.  

%


\subsection{Multimodal Debiasing}
Multimodal debiasing focuses on mitigating the over-reliance on unimodal priors, which often leads to deceptive shortcuts and hinders cross-modal reasoning \cite{Agrawal2018}. 
In VG tasks, language bias is particularly prominent, where models are often influenced by statistical correlations in queries rather than being guided by visual evidence \cite{UnveilingInternalReasoningModes,QuestionConditionedDebiasing}. 

Existing debiasing strategies primarily fall into three categories. 
The first involves data-level interventions, such as synthesizing counterfactual samples \cite{CSS} or utilizing text filtering \cite{InstanceVG} to balance the priors' distribution. However, these methods may discard critical details or introduce artificial noise. 
The second category builds upon architecture-based suppression, which designs specialized gating mechanisms or auxiliary branches to capture and subtract bias-related signals \cite{AutomaticallyNeutralizingSubjectiveBiasInText}. 
The third category  centers on adaptive recalibration and task-specific debiasing, which dynamically modulates feature interactions or purifies biased signals based on contextual intent to ensure that the reasoning process remains grounded in genuine cross-modal evidence \cite{GBT,USRI-REID}. 
In contrast, BARE proposes a complementary debiasing framework that prioritizes task-relevant linguistic cues and rectifies perceptual biases to suppress linguistic shortcuts and enhance referential representations, improving robustness and precision.

\subsection{Vision-Language Pretraining}
From the perspective of feature encoding, existing vision-language pretraining (VLP) models can be broadly categorized into two paradigms: one-tower and dual-tower architectures.
One-tower models \cite{vilbert,BEiT,ViLT,One-peace} process image and text inputs within a shared encoder by concatenating their embeddings, thereby facilitating cross-modal interactions
throughout the feature extraction process.
In contrast, dual-tower models \cite{CLIP} employ separate encoders for each modality, processing the unmerged inputs independently and relying on cross-modal interactions for alignment. 
%
In this paper, we adopt a one-tower paradigm that advances multimodal feature fusion from the late fusion stage into the early backbone stage.

\subsection{Low-Rank Adaptation}
As one of the most widely adopted methods in Parameter-Efficient Fine-Tuning (PEFT), Low-Rank Adaptation (LoRA) \cite{LoRA} introduces low-rank matrices to explicitly represent weight modifications while keeping the original parameters frozen. 
These explicit changes can subsequently be merged with the original weight priors during inference, ensuring that no additional inference latency is incurred compared to vanilla models. 
Existing works primarily employ LoRA for fine-tuning large language models (\textit{e.g.}, LLaMA \cite{LLaMA2} ), with only a limited number exploring its application in cross-modal tasks (\textit{e.g.}, CLIP-LoRA \cite{CLIP-LoRA} and M2IST \cite{M2IST}). 
{In this paper, we leverage LoRA for efficient adaptation, enabling effective bias suppression and referential reasoning enhancement for grounding tasks.}

%% file: Body/3-method.tex
In this section, we detail the architecture and technical implementation of our BARE framework. We first provide a high-level overview of the framework pipeline, followed by an in-depth discussion of the three technical modules. Finally, we elaborate on the training objectives and optimization strategies.
\subsection{Framework Overview}
\begin{figure*}[htbp]\centering
        \setlength{\belowcaptionskip}{-8pt}
	\includegraphics[scale=0.618]{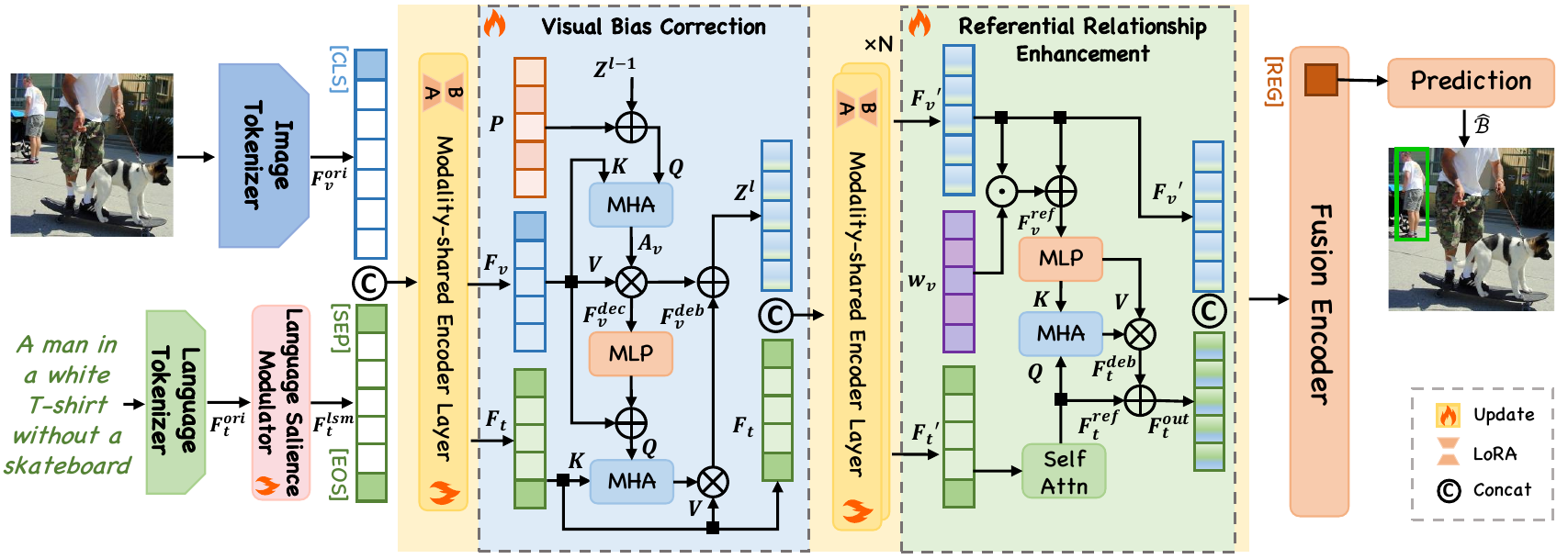}

	\caption{Schematic overview of the BARE framework. The model extracts visual and textual features via dedicated tokenizers, with linguistic signals refined by the Language Salience Modulator (LSM). These features are then integrated within a modality-shared encoder that incorporates Visual Bias Correction (VBC) and Referential Relationship Enhancement (R$^2$E) modules to suppress modality-specific biases and reinforce compositional reasoning. Finally, a fusion encoder leverages a regression token ([REG]) to aggregate multimodal information for precise coordinate prediction.
                }
    \label{FIG_3}
\end{figure*}

%
Given an image and a query, the VG model needs to predict a bounding box $\mathcal{B}=\{x,y,w,h\}$ that accurately surrounds the target described by the query.
As shown in Fig. \ref{FIG_3}, our BARE framework consists of four components as follows.

\begin{figure}[t]\centering
	\includegraphics[width=6cm]{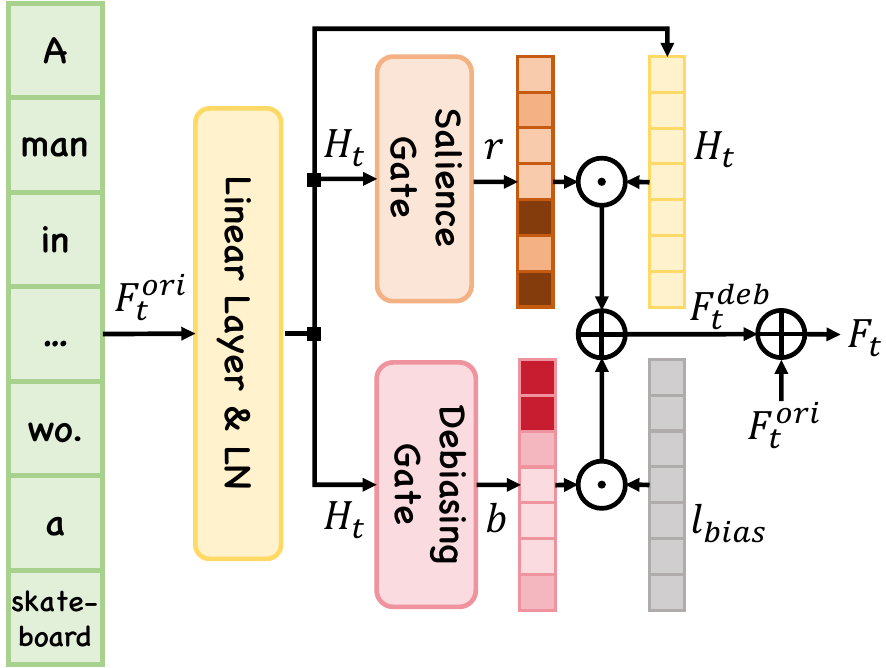}

    \caption{Architecture of the Language Salience Modulator (LSM). The module leverages a dual-gating mechanism to refine language signals: a salience gate accentuates task-critical cues, while a debiasing gate suppresses deceptive linguistic shortcuts via a learnable shared prior $l_{\text{bias}}$. The final representation is obtained by interpolating the original and debiased features.}
    \vspace{-8pt}
    \label{FIG_4_LSM}
  
\end{figure}

\noindent\textbf{Tokenizer.}
The input consists of an image $\mathcal{I} \in \mathbb{R}^{H_0 \times W_0 \times 3}$ and a corresponding query description $\mathcal{T} \in \mathbb{R}^{L}$, where $H_0$, $W_0$ denote the image size and $L$ is the number of words in the text.
{We extract visual embeddings using a convolutional tokenizer, which projects the input image into a sequence of tokens $F_v^{ori} \in \mathbb{R}^{N \times C}$, where $N = HW$ and $C$ is the hidden dimension. Here, $H$ and $W$ are downsampled to $\frac{1}{16}$ of the original $H_0$ and $W_0$.}
%
The input text is tokenized using a SentencePiece tokenizer \cite{sentencepiece} with a vocabulary size of 64,010 and then projected into word embeddings.
Following standard practice \cite{attention_is_all_you_need}, we prepend a learnable [SEP] token and append an [EOS] token to form the final text embeddings $F_t^{ori} \in \mathbb{R}^{M \times C}$, where $M$ is the sequence length after tokenization. 
Given that the pre-interaction text embeddings retain substantial original characteristics, we introduce a \textit{Language Salience Modulator} (LSM) to adjust them at the token level, resulting in debiased text embedding $F_t^{lsm} \in \mathbb{R}^{M \times C}$.

\noindent\textbf{Modality-shared Encoder.}
We first concatenate the visual and textual embeddings along the sequence dimension to form the joint multimodal features $X = \text{concat}[F_v^{ori}; F_t^{lsm}]$. 
The encoder comprises stacked layers of self-attention and feed-forward networks (\textit{i.e.} modality experts) to process the fused inputs.  
To enhance parameter efficiency, we integrate Low-Rank Adaptation (LoRA) into the encoder, enabling lightweight task-specific tuning. 
Building upon this, we further introduce two bias-aware and reasoning-enhanced modules: \textit{Visual Bias Correction} (VBC) and \textit{Referential Relationship Enhancement} (R$^2$E).
These modules enhance modality-specific representations and referential reasoning by leveraging cross-modal guidance and effectively incorporate low-rank matrices for gradient updates.
Both modules are seamlessly integrated into the encoder and preserve the original embedding dimensions, thus the final encoder output $X^* \in \mathbb{R}^{(N+M) \times C}$ maintains the same shape as the input $X$.
%

\noindent\textbf{Fusion Encoder.}
Serving as the central fusion component, the encoder progressively integrates and refines multimodal representations via a deep stack of  Transformer layers. 
It takes the encoder output $X^*$ as input and prepends a learnable regression token [REG] to obtain the augmented features $X_g \in \mathbb{R}^{(N+M+1) \times C}$.
Finally, the [REG] token interacts with refined multimodal features, guiding the decoder to focus on relevant cross-modal cues for accurate grounding.  

\noindent\textbf{Prediction Head.}
The primary goal of this module is to use the [REG] token from $X_g$ to predict the bounding box $\hat{\mathcal{B}}$, with supervision from the ground truth $b$. 
To this end, we optimize the model using the smooth L1 loss \cite{fast-rcnn} and the GIoU loss \cite{giou}. The training objective is formulated as follows:
\begin{equation}
    \mathcal{L}_{coor}=\lambda_{l_1}\mathcal{L}_{smooth-l1}(\mathcal{B},\hat{\mathcal{B}})+\lambda_{giou}\mathcal{L}_{giou}(\mathcal{B},\hat{\mathcal{B}}).
\end{equation}

\subsection{Language Salience Modulator}
Due to frequency-based priors in language, high-frequency queries (\textit{e.g.}, \textit{``a man in"} or \textit{``a dog with"}) tend to induce deceptive shortcuts \cite{DeceptiveSemanticShortcuts,UnveilingInternalReasoningModes} in the model, leading it to rely on pattern matching or corpus statistics to predict targets instead of carefully reasoning about the underlying semantics of the expression. 
Generation-based language debiasing methods \cite{AutomaticallyNeutralizingSubjectiveBiasInText,ReinforcedSequenceTrainingBasedSubjectiveBiasCorrection} primarily address subjective biases in phrasing thus fail to strengthen referential cues, while text filtering \cite{InstanceVG} may discard critical details. To overcome these limitations, as shown in Fig. \ref{FIG_4_LSM}, we introduce a Language Salience Modulator (LSM) that dynamically reweights language features for fine-grained modulation of linguistic signals.

Specifically, given the text embeddings $F_t^{ori} \in \mathbb{R}^{M \times C}$ produced by the tokenizer, the LSM module first applies a linear projection to obtain an intermediate representation $H_t$. 
\begin{equation}
    H_t = \text{LN}(F_t^{ori} W_h),
\end{equation}
where $W_h$ is the projection matrix and $\text{LN}$ is the layer normalization operation. 
Subsequently, we introduce two gating mechanisms operating from complementary perspectives: a salience gate that amplifies referential signals and a debiasing gate that suppresses distracting components.
\begin{equation}
\begin{aligned}
    r &= sigmoid(H_t w_r),  \\
    b &= sigmoid(H_t w_b),  \\
\end{aligned}
\end{equation}
where $w_r,w_b$ are the parameters of the two gate mechanisms. 
Then, we fuse the features from both gates to obtain debiased representations by:
\begin{equation}
    F_t^{deb} = r \odot H_t - \lambda \, b \odot l_{\text{bias}},
\end{equation}
where $\lambda$ is a scaling factor, and $\odot$ is the element-wise multiplication operation. 
Here, $l_{\text{bias}} \in \mathbb{R}^{M \times C}$ denotes a randomly initialized, learnable shared prior that captures linguistic patterns commonly observed. Although high-frequency queries may differ in expression, previous studies \cite{GVQA,RUBi,PAR,Coop} indicate that they tend to cluster along similar directions in the joint embedding space. 
A straightforward solution is to reweight the original text features \cite{ZS-TAD,PBAL,TTE}, but this makes the bias pattern input-dependent and often leads to over-correction at the token level. 
In contrast, $l_{\text{bias}}$ mitigates overfitting to token-specific bias and drives the model to capture a shared bias direction that can be more effectively removed.
Finally, to smoothly interpolate between the original and debiased representations, we leverage convex combination \cite{DeepFeatureInterpolation} to obtain the final output:
\begin{equation}    
    {F}_t^{lsm} = F_t^{ori} + \alpha \left(F_t^{deb} - F_t^{ori}\right),
\end{equation}
where $\alpha$ is a learnable scaling factor between 0 and 1.

\subsection{Visual Bias Correction}
\label{sec_3_2}

While one-tower frameworks attempt to bridge the modality gap \cite{MindTheGap} inherently present in dual-tower architectures, they still suffer from over-entangled multimodal representations. 
{To address this issue, we propose VBC to mitigate the blending of modality-specific signals by decoupling entrenched visual priors that often lead to deceptive associations. }
Inspired by studies that visual representations across layers capture semantics at varying levels of abstraction \cite{BehindTheScene,AnalyzingTRM}, we apply VBC across multiple layers to facilitate multi-granularity decoupling and interaction.

Given the split visual features $F_v\in \mathbb{R}^{N\times C}$ and textual features $F_t \in \mathbb{R}^{M\times C}$, VBC leverages semantic prototypes to extract salience information and disentangle latent visual biases. 
The semantic prototypes $P \in \mathbb{R}^{K \times C}$ are designed as a set of $K$ learnable tokens, which are randomly initialized and share same shape with the visual tokens.
As illustrated in Fig. \ref{FIG_3}, in the vision debiasing stage, the modality-shared encoder consists of $L$ layers of VBC. At layer $l$, the output of the previous layer, $Z^{l-1} \in \mathbb{R}^{N \times C}$, is combined with the semantic prototypes $P$ to attend to local visual features $F_v$ with multi-head attention (MHA), resulting in the attention matrix $A_v$. This processing can be formalized by:
\begin{equation}    
    Q=(Z^{l-1}+P)W_q,\quad K=F_v W_k.
\end{equation}

The resulting attention map $A_v=softmax({QK^\top}/{\sqrt{d}})$ capture spatial correspondences, with each prototype token attending to a specific  region, enabling the model to perceive diverse visual regions that may implicitly encode visual biases and subsequently mitigate their influence. 
These maps are then used to compute the decoupled region-level features via:
\begin{equation}
F_v^{dec}=A_{v}(F_v W_{v})^\top.
\end{equation}

While the decoupled region features $F_{v}^{dec}$ effectively alleviate deceptive visual biases, the lack of explicit alignment with the language modality still hampers cross-modal grounding. 
{To this end, we introduce a reasoning enhancement strategy that integrates language cues into the visual stream, thereby enhancing its referential perception.} 
Specifically, the debiased text features $F_t^{lsm}$ have already captured rich contextual dependencies through the encoder, providing semantically coherent representations. 
We therefore employ MHA with MLP-projected visual features as $Q$, and these linguistically enriched features $F_t$ as $K$ and $V$, producing enhanced visual representations that explicitly align regions with referring expressions as: $F_v^{deb}={softmax}\!\left({QK^\top}/{\sqrt{d}}\right)V$. 
%

Finally, we fuse the two visual features through residual addition: $Z^l = F_{v}^{dec} + F_{v}^{deb}$, and concatenate the fused visual features with the original language features to obtain the final VBC output as  $X_{r}=concat[Z^l,F_t]$. 
{In this manner, each visual region is adaptively debiased by language cues, resulting in more precise and flexible grounding that effectively overcomes the limitations of static patch-based representations.}

\subsection{Referential Relationship Enhancement}
Although recent approaches \cite{HiVG,SimVG} have achieved advanced grounding performance, yet addressing complex expressions with multiple attributes and inter-object relations remains challenging, often requiring explicit referential and compositional reasoning \cite{vg_survey,RVG-Tree,VLTVG}.
Motivated by these observations, we design the R$^2$E module to enhance the model's capacity for referential reasoning. Applied at the final encoder layer, R$^2$E leverages global context to refine referential cues, guiding more accurate and semantically consistent grounding.

Specifically, similarly to VBC, the R\textsuperscript{2}E module operates on the bias-aware visual features $F_v' \in \mathbb{R}^{N\times C}$ and language features $F_t'\in \mathbb{R}^{M\times C}$. 
Unlike VBC query-driven visual decoupling, R$^2$E focuses on the linguistic stream, adaptively modulating referential semantics. 
To this end, we first apply a learnable feature-wise weight matrix $w_v \in \mathbb{R}^{M \times C}$ to enhance informative visual tokens and suppress irrelevant ones.
\begin{equation}
    F_{v}^{ref}=w_v \odot F_v'+F_v',
\end{equation}
this residual connection helps maintain feature stability while enhancing the representation of target-aware tokens. 
Subsequently, we apply an MLP to project and enrich the visual features, mapping them into an attention-consistent space. 
Meanwhile, self-attention is applied over $F_t'$ to model inter-word dependencies, resulting in language-aware features via $F_{t}^{ref}=SelfAttn(F_t')$. 
Finally, we use the context-aware text features $F_t^{ref}$ (with token dependencies already modeled) as $Q$, and the transformed visual features as the $K$ and $V$ in MHA,  to obtain the debiased text features as $F_t^{deb}={softmax}\!\left({QK^\top}/{\sqrt{d}}\right)V$.

The two branches are then fused via residual addition as $F_t^{out} = F_{t}^{ref} + F_{t}^{deb}$, and the final R\textsuperscript{2}E output is obtained by concatenating the updated language features with the visual ones as $X^{*} = concat[F_v',F_t^{out}]$. 
Guided by visual evidence, R\textsuperscript{2}E explicitly models fine-grained semantic dependencies between refined linguistic cues and visual entities, thereby strengthening structured referential reasoning and improving robustness under complex or ambiguous expressions.



\input{tables/table1_comp_sota.tex}

\subsection{Training Objectives}

\noindent\textbf{Image-Text Alignment.} 
To establish correspondences between images and text while ensuring stable training, we use Image-Text Alignment (ITA) \cite{CLIP} as a foundational supervision objective. 
Taking text as the anchor, the text-to-image contrastive loss is defined as: 
\begin{equation}
    \mathcal{L}_{t2i}=-\frac{1}{B}\sum_{i}^{B} \log \frac{exp(sim<{t_i}^{\top},v_i>/\tau)}{\sum_{j=1}^{B}exp(sim<{t_i}^\top,v_j>/\tau)},
\end{equation}
{where $\textit{sim}<\! \cdot\! >$ denotes cosine similarity and $\tau$ is a temperature factor.}
The image-to-text loss is computed analogously. The overall ITA loss combines both formulations as:   
\begin{equation}
    \mathcal{L}_{ITA}=\frac{1}{2} (\mathcal{L}_{i2t}+\mathcal{L}_{t2i}).
\end{equation}
%

\noindent\textbf{Region-Text Alignment.} 
To capture region-level constraints, we {leverage} {Region-Text Alignment (RTA) \cite{HiVG}} to explicitly associate visual patches with the textual queries. 
Specifically, we extract the text-aggregated [EOS] token, denoted as the $t_{eos}$. After normalization, we compute its similarity with each image patch $v_i^j$: 
\begin{equation}
    {s}_{i}^j=\sigma (<{t_{eos}}^\top,v_i^j>),j=1,2,\dots,N,
\end{equation}
where $\sigma$ denotes the sigmoid function and $N$ represents the number of patch tokens in the downsampled feature maps. 
Finally, we optimize the probabilities \rev{$\boldsymbol{s_i}=(s_i^1,\dots,s_i^{N})$} using a combination of Focal Loss \cite{focal} and Dice Loss \cite{dice}: 
\begin{equation}
    \mathcal{L}_{RTA}=\frac{1}{B}\sum_{i=1}^B(\lambda_{f}\mathcal{L}_{f}(\boldsymbol{s_i},\boldsymbol{g_i})+\lambda_{d}\mathcal{L}_{d}(\boldsymbol{s_i},\boldsymbol{g_i})),
\end{equation}
where $\lambda_f$ and $\lambda_d$ are coefficients, and $\boldsymbol{g_i} \in \mathbb{R}^{1\times H \times W}$ is the downsampled box mask, where $H=\frac{H_0}{16}$ and $W=\frac{W_0}{16}$. 
%

\noindent\textbf{Pixel-Text Alignment.} 
We adopt Pixel-Text Alignment (PTA) \cite{SegVG} for fine-grained multimodal alignment with full use of bounding box annotations.
{Specifically, we upsample the visual features to the original image resolution and leverage the bounding box annotations by treating them as segmentation masks. }
The prediction map \rev{$\boldsymbol{m_v}$} is then optimized against the ground truth mask \rev{$\boldsymbol{m_g}$} as: 
\begin{equation}
    \mathcal{L}_{PTA}=\frac{1}{B}\sum_{i=1}^B(\lambda_{f}\mathcal{L}_{f}(m_v^i,m_g^i)+\lambda_{d}\mathcal{L}_{d}(m_v^i,m_g^i)).
\end{equation}

\noindent\textbf{Training Loss.}
The alignment loss can be defined as:
\begin{equation}
    \mathcal{L}_{align}=\mathcal{L}_{ITA}+
    \mathcal{L}_{RTA}+\mathcal{L}_{PTA}.
\end{equation}

With the prediction head locating region coordinates $b$, the final training objective can be formulated as:
\begin{equation}
    \mathcal{L}=\mathcal{L}_{coor}+
    \mathcal{L}_{align}.
\end{equation}

%% file: tables/table1_comp_sota.tex
\begin{table*}[t]
\centering
\caption{
Performance comparison (\%) with SOTA methods on RefCOCO/+/g, ReferIt and Flickr30k. \underline{Underline} highlights the best performance among base models, while \textbf{bold} denotes the best among large models. The term \textit{w/} denotes with.
}
\resizebox{1 \textwidth}{!}{%
\small
\begin{tabular}{c|c|c|ccc|ccc|cc|c|c}
\toprule
\multirow{2}{*}{Models} & \multirow{2}{*}{Venue}    & {Visual} & \multicolumn{3}{c|}{RefCOCO}                   & \multicolumn{3}{c|}{RefCOCO+}                  & \multicolumn{2}{c|}{RefCOCOg}                      & ReferIt &Flickr30k  \\
                        &                           & Encoder  & \textit{val} & \textit{testA} & \textit{testB} & \textit{val} & \textit{testA} & \textit{testB} & \textit{val-u} & \textit{test-u} & \textit{test}   &\textit{test}\\ 
\midrule

\multicolumn{13}{l}{\textbf {(1) Fine-tuning setting \textit{w/} pretrained close-set detection and language model on single dataset}} \\

QRNet \cite{QRNet}        &CVPR'22   & Swin-S \cite{SwinTransformer}   & 84.01     & 85.85     & 82.34     & 72.94     & 76.17     & 63.81     & 73.03     & 72.52 & 74.61   & 81.95   \\
VLTVG \cite{VLTVG}        &CVPR'22   & RN101 \cite{ResNet}    & 84.77     & 87.24     & 80.49     & 74.19     & 78.93     & 65.17     & 76.04     & 74.18 & 71.98   & 79.84   \\
VG-LAW \cite{VG-LAW}      &CVPR'23   & ViT-B \cite{CLIP}    & 86.06     & 88.56     & 82.87     & 75.74     & 80.32     & 66.69     & 76.90     & 76.96 & 77.22   &  --   \\
MFSD \cite{MFSD}           &TCSVT'24  & RN101 \cite{ResNet}    & 86.82     & 88.75     & 82.60     & 76.22     & 80.75     & 67.33     & 77.86     & 76.24 &  74.03  & 81.45   \\
SegVG \cite{SegVG}       &ECCV'24   & Swin-S \cite{SwinTransformer}   & 86.84     & 89.46     & 83.07     & 77.18     & 82.63     & 67.59     & 78.35     & 77.42 & 75.59   &  --  \\
\midrule

\multicolumn{13}{l}{\textbf {(2) Fine-tuning setting \textit{w/} pretrained vision-language model on single dataset}} \\
CLIP-VG \cite{CLIP-VG}     & TMM'23   & CLIP-B \cite{CLIP}   & 84.29     & 87.76     & 78.43     &69.55      & 77.33      & 57.62    & 73.18     & 72.54 & 70.89   & 81.99   \\
DUET \cite{VisualGroundingWithDualKnowledgeDistillation}  &TCSVT'24  & Swin-S \cite{SwinTransformer}    & 87.27     & 89.70     & 83.37     & 75.86     & 80.00     & 67.65     & 78.47     & 77.66 &  73.05  & 81.35   \\
HiVG-B \cite{HiVG}      & ACM MM'24 & CLIP-B \cite{CLIP}   & 87.32     & 89.86     & 83.27     &78.06      & 83.81      &68.11     & 78.29     & 78.79 & 75.22   &  82.11  \\
HiVG-L \cite{HiVG}       & ACM MM'24 & CLIP-L \cite{CLIP}   & 88.14     & 91.09     & 83.71     &80.10      & 86.77      &70.53     & 80.78     & 80.25 &  76.23  & 82.16   \\
SimVG-B \cite{SimVG}      & NeurIPS'24  & BEiT-B \cite{BEiT}  & 87.63     & 90.22     & 84.04     &78.65      & 83.36      &71.82     & 80.37     & 80.51 & 74.83   & 82.04   \\
SimVG-L \cite{SimVG}      & NeurIPS'24  & BEiT-L \cite{BEiT} & 90.61     & 92.53     & 87.68     &85.36      & 89.61      &79.74     & 85.99     & {86.83}  & 79.30   &  82.61  \\    
PLVL \cite{PLVL}   & arXiv'25   & ViT-B \cite{CLIP}  & \underline{89.02}     & 90.21     & \underline{86.72}     & 79.19     & 84.97     & 71.55     & 81.58     & 81.16 &  --   &  --  \\
\midrule
\rowcolor{gray!10}
BARE-B (Ours)        & --   & BEiT-B \cite{BEiT}    & {88.69}     & \underline{91.61}    & {84.79}     &\underline{81.29}      & \underline{86.78}      &\underline{74.75}     & \underline{82.73}     & \underline{83.17} & \underline{75.32}   &  \underline{82.28}  \\
\rowcolor{gray!10}
BARE-L (Ours)        & --   & BEiT-L \cite{BEiT}    & \textbf{92.37}     & \textbf{93.77}     & \textbf{89.21}     &\textbf{87.85}      & \textbf{91.78}      &\textbf{83.17}     & \textbf{88.75}     & \textbf{88.87} &  \textbf{79.98}  & \textbf{83.34}   \\
\midrule
\midrule

\multicolumn{13}{l}{\textbf {(3) Fine-tuning setting \textit{w/} pretrained open-set detection model on mixed datasets}} \\
YORO \cite{YORO}        & ECCV'22  & ViLT \cite{ViLT}     & 82.90     & 85.60     & 77.40     & 73.50     & 78.60      & 64.90    & 73.40     & 74.30 & 71.90   & --  \\
UniTAB \cite{UniTAB}      & ECCV'22  & RN101  \cite{ResNet}   & 86.32     & 88.84     & 80.61     & 78.70     & 83.22      & 69.48    & 79.96     & 79.97 & --      &  79.38 \\
OFA-B \cite{OFA}       & ICML'22  & OFA-B \cite{OFA}     & 88.48     & 90.67     &83.30      & 81.39     & {87.15}    & 74.29 & 82.29& 82.31 & --      & -- \\
OFA-L  \cite{OFA}      & ICML'22  & OFA-L \cite{OFA}     & 90.05     & 92.93     & 85.26     & 85.80     & 89.87      & 79.22    & 85.89     & 86.55 & --   &  -- \\
DQ-DETR  \cite{DQ-DETR}    & AAAI'23  & RN101 \cite{ResNet}     & 88.63     & 91.04     & 83.51     & 81.66     & 86.15      & 73.21    & 82.76     & 83.44 & --      &  --  \\
Grounding-DINO \cite{Grounding-DINO} & ECCV'24  & Swin-T \cite{SwinTransformer}  & 89.19     & \underline{91.86} & 85.99  & 81.09     & \underline{87.40}      & 74.71    & 84.15     & 84.94 & --      & --  \\

\midrule
\rowcolor{gray!10}
BARE-B (Ours)        & --   & BEiT-B \cite{BEiT}    & \underline{90.28}     & {91.54}    & \underline{87.93}     &\underline{83.14}      & 87.06      &\underline{77.25}     & \underline{85.37}     & \underline{86.03} & \underline{77.16}   &  \underline{82.45}  \\
\rowcolor{gray!10}
BARE-L (Ours)        & --   & BEiT-L \cite{BEiT}    & \textbf{92.83}     & \textbf{94.28}     & \textbf{90.78}     &\textbf{88.36}      & \textbf{91.03}      &\textbf{84.19}     & \textbf{90.58}     & \textbf{90.65} &  \textbf{80.58}  & \textbf{83.68}   \\

\bottomrule
\end{tabular}%
}
\label{table1}
\vspace{-8pt}

\end{table*}

%% file: Body/4-Experiments.tex
\subsection{Dataset and Evaluation}
We evaluate our model on five benchmarks: three for REC (RefCOCO/+/g \cite{Refcoco,Refcocog}) and two for Phrase Grounding (PG) (ReferIt \cite{ReferIt} and Flickr30K Entities \cite{flickr30k}).
In RefCOCO/+, the \textit{testA} split contains only person-related queries, while \textit{testB} focuses on other object categories.
RefCOCOg features significantly longer and more descriptive expressions, whereas queries in PG are typically short noun phrases.
Following prior work, TransVG \cite{TransVG}, we adopt Intersection-over-Union (IoU) as the evaluation metric: a prediction is deemed correct if its IoU with the ground-truth box is at least 0.5.
We report accuracy as the final performance metric across all datasets.

\begin{figure}[t]\centering
	\includegraphics[width=8.8cm]{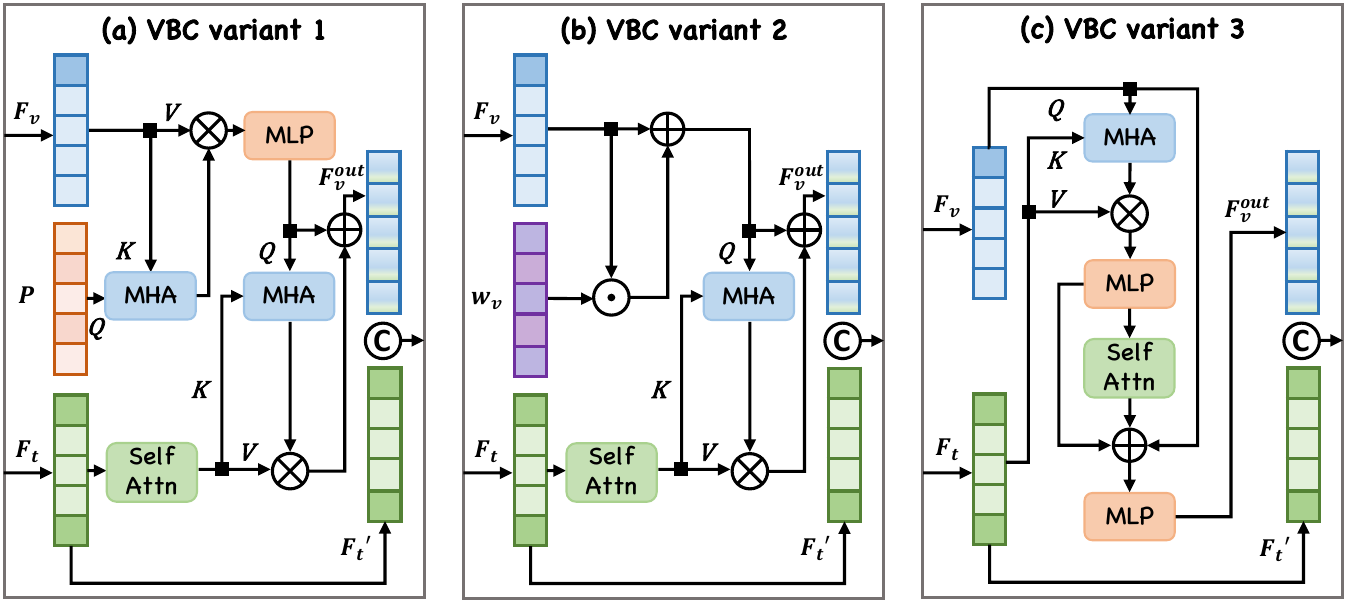}

	\caption{Detailed architectural variants of the proposed VBC modules. 
    (a) The variant employing an alternative MHA between the semantic prototypes and features. 
    (b) The variant that replaces MHA with a dot-product interaction parameterized by learnable weights.
    (c) The variant configured without semantic prototypes, utilizing an interaction scheme for feature processing.}    
    \label{FIG_5_VBC_variant}
    \vspace{-8pt}
\end{figure}

\input{tables/table2_comp_cost}

\subsection{Implementation Details} 
\noindent\textbf{Network Architecture.}
We adopt BEiT-3 \cite{BEiT} as the backbone, using its base and large variants for BARE-B (default) and BARE-L, respectively.
BARE-B comprises 12 encoder layers. The visual bias correction module is inserted every 4 layers, specifically after layers 1, 5, and 9, while the referential enhancement module is applied only at the final encoder layer.
We employ a post-layer normalization design, with a hidden dimension of 768 and a feed-forward dimension of 3072 (\textit{i.e.}, $4 \times 768$).
In BARE-L, the decoder projection dimension is increased to 1024, and the feed-forward dimension is expanded to 4096, with all other settings consistent with BARE-B. 

\noindent\textbf{Training Details.}
We follow the second and third experimental settings in Tab.~\ref{table1}.
Input images are resized to $224 \times 224$ with augmentations following \cite{HiVG}, while queries are tokenized to a maximum length of 77.
To maintain efficiency and stability, the BEiT-3 backbone remains frozen. We apply LoRA to the encoder's Q, K, V, and feed-forward matrices. In contrast, task-specific modules, including LSM, VBC, and R$^2$E, are trained directly without low-rank constraints.
In the LSM module, $\alpha$ and $\lambda$ are initialized to 0.5 and 0.1, respectively, and are updated as training progresses.
The model is trained for 60 epochs using the AdamW \cite{CLIP-VG} optimizer  with a batch size of 32, and both learning rate and weight decay set to $10^{-4}$.

\noindent\textbf{Inference Details.} 
During the inference stage, all model parameters, including the backbone and the task-specific components, remain fixed. The prediction process follows the same pipeline as training, where the regression token directly yields the bounding box coordinates without any additional post-processing, ensuring high inference efficiency.

\begin{figure}[t]\centering
	\includegraphics[width=8.6cm]{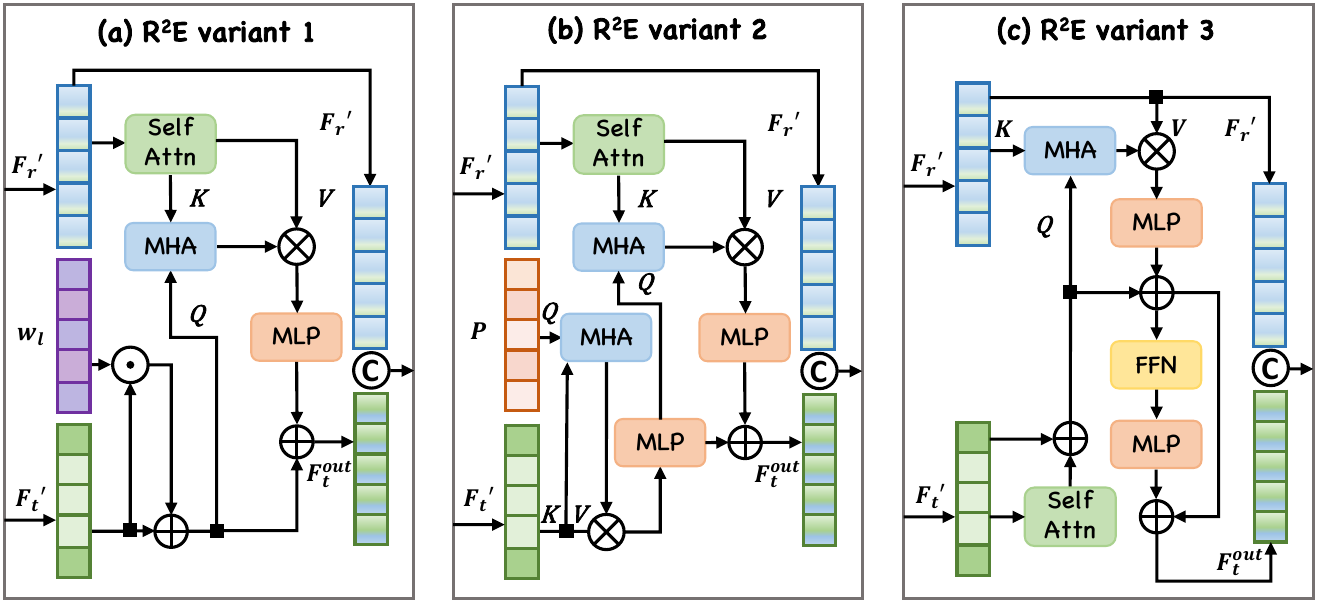}

\caption{Detailed architectural variants of the proposed R$^2$E module. 
    (a) The variant that directly applies the learnable weights to the linguistic features.
    (b) The variant that removes the weighting mechanism and focuses on text refinement via auxiliary learnable queries.
    (c) The variant that replaces the weighting operation with the MHA approach. }    
    \label{FIG_5_R2E_variant}
    \vspace{-8pt}
\end{figure}

\input{tables/table3_main_ablation}


\begin{figure*}[htbp]\centering
	\includegraphics[width=1.0\textwidth]{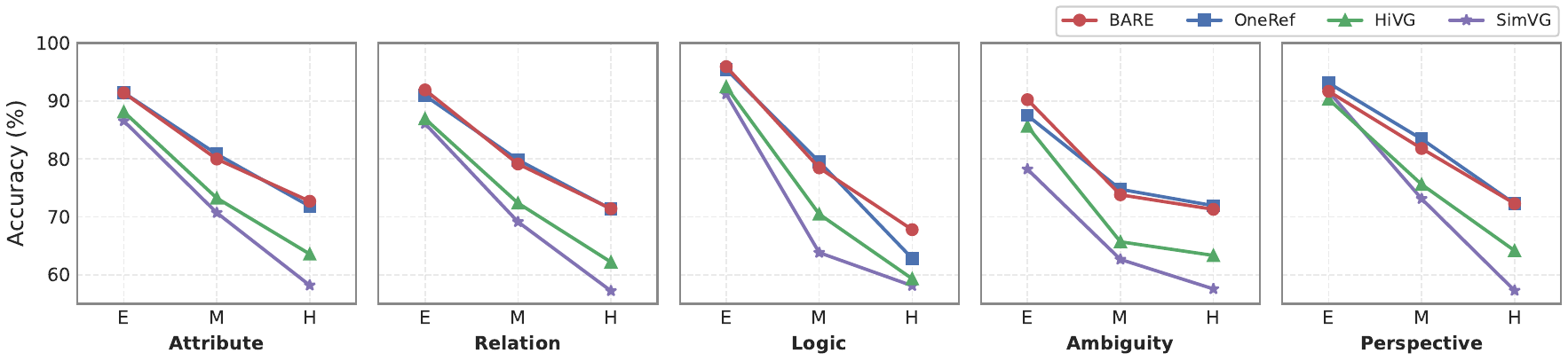}
	\vspace{-20pt}
	\caption{Fine-grained comparison of grounding accuracy across five referential categories and three difficulty levels (E, M, and H denote Easy, Medium, and Hard, respectively) on the RefCOCOg test set. We compare BARE with HiVG \cite{HiVG}, SimVG \cite{SimVG}, and OneRef \cite{oneref}. Results for SimVG are reproduced by us under the same experimental settings for fair comparison. The details of five referential categories  are elaborated in Template~\ref{table_prompt_tpl}. }
    \label{FIG_6}
    \vspace{-8pt}
\end{figure*}

\subsection{Comparison with SOTA Methods} 
\noindent\textbf{Model Performance.}
We compare BARE with state-of-the-art methods on RefCOCO/+/g \cite{Refcoco,Refcocog-umd}, ReferIt \cite{ReferIt}, and Flickr30k \cite{flickr30k} under three representative settings, as summarized in Tab.~\ref{table1}. 
(1) Fine-tuning with close-set detection and language backbones yields competitive results (\textit{e.g.}, SegVG \cite{SegVG}), but it is generally outperformed by recent methods built upon pretrained vision-language models. 
(2) Fine-tuning with pretrained vision-language models on a single dataset: BARE-B achieves the best base-model performance across all RefCOCO/+/g splits (\textit{e.g.}, 84.79 on RefCOCO testB, 74.75 on RefCOCO+ testB, and 83.17 on RefCOCOg test-u), and surpasses SimVG-B by 0.75\%, 2.93\%, and 2.66\% on the corresponding splits. BARE-L further sets the strongest overall performance among large models, reaching 93.77/89.21 on RefCOCO testA/testB and 79.98/83.34 on ReferIt/Flickr30k.
(3) Fine-tuning with data-mixed pretraining: BARE remains strong under this setting---BARE-B improves over Grounding DINO by 1.94\%, 2.54\%, and 1.22\% on RefCOCO/+/g (testB/testB/val-u), while BARE-L outperforms OFA-L by 5.52\%, 4.97\%, and 4.69\% on the same splits.

\input{tables/table4_vbc_ablation}

\noindent\textbf{Computational Cost.}
As shown in Tab.~\ref{table_comp_cost}, BARE updates only 4.7M parameters, accounting for just 1.7\% of total model size.
Despite updating far fewer parameters---only 22.4\% of CLIP-VG and 11.5\% of HiVG---BARE attains the best accuracy on RefCOCO testA (91.6\%). Notably, our method does \emph{not} target the fastest inference (11.08 FPS / 255s versus 14.67 FPS / 192s for CLIP-VG \cite{CLIP-VG}), but prioritizes achieving superior accuracy under an exceptionally lightweight fine-tuning paradigm.
Although the BEiT-B backbone incurs higher FLOPs than CLIP-B due to its larger hidden dimension, BARE still offers a favorable accuracy--efficiency trade-off overall. 
These results suggest that lightweight fine-tuning, when guided by principled designs such as LSM, VBC, and R$^2$E, can match or even outperform full fine-tuning while updating orders of magnitude fewer parameters.


\subsection{Ablation Studies} 

\noindent\textbf{Ablation Study of the Main Modules.} 
Tab.~\ref{table_main_ablation} systematically evaluates the individual contributions of LSM, VBC, R$^2$E, and the Fusion Encoder. 
The results indicate that each module plays a distinct and vital role in the framework. 
Specifically, while all individual modules provide clear improvements, the VBC module emerges as the most effective single-module component (82.42\% on test), highlighting its strength in correcting visual bias. 
The full configuration yields the best overall performance (83.17\% on test), which validates the functional complementarity among LSM, VBC, and R$^2$E in suppressing deceptive shortcuts and reinforcing referential reasoning. 
Furthermore, removing the fusion encoder results in a dramatic performance drop of 3.79\%, underscoring its indispensable role in aggregating refined multimodal features.

\input{tables/table5_lsm_r2e.tex}

\begin{figure*}[htbp]\centering
	\includegraphics[scale=0.675]{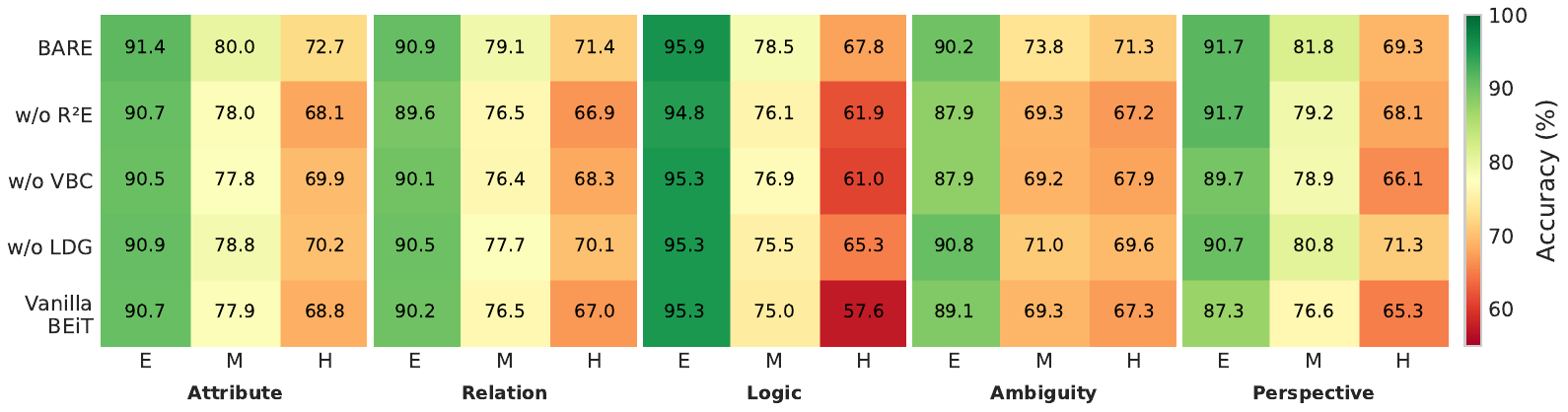}
    \vspace{-20pt}
	\caption{Fine-grained heatmap visualization of grounding accuracy across five referential types and three difficulty levels (E, M, and H denote Easy, Medium, and Hard, respectively). Each row corresponds to a model variant: BARE (full model), its key ablations (w/o VBC, w/o R$^2$E, w/o LSM), and the vanilla BEiT baseline.   The referential type definitions are elaborated in Template~\ref{table_prompt_tpl}.}
    \vspace{-8pt}
        
    \label{FIG_7}
\end{figure*}

\noindent\textbf{Ablation Study of LSM Module.} 
Tab.~\ref{table_lsm_r2e} investigates the impact of the dual-gating mechanism within the LSM module. 
We observe that removing either the salience gate or the debiasing gate leads to a clear performance drop. 
Specifically, the exclusion of the debiasing gate results in a substantial decline (0.95\%), which confirms that suppressing high-frequency linguistic shortcuts is critical for mitigating co-occurrence biases. 
Similarly, the salience gate further boosts accuracy by accentuating target-relevant cues. 
The full LSM's optimal performance validates the synergistic effect of both gates in purifying linguistic representations.

\noindent\textbf{Ablation Study of VBC Module.} 
Tab.~\ref{table_vbc_ablation} evaluates the structural design and key components of the VBC module. 
We first explore the impact of stride size and insertion depth, motivated by the varying levels of semantic abstraction across encoder layers. 
Empirical results show that a stride of 4 and an early insertion (Layer 1$\sim$2) yield the optimal performance, while deeper insertions lead to a gradual decline. 
Regarding prototype initialization, randomly initialized learnable tokens outperform those derived from visual features (\textit{e.g.}, via mean or max pooling), suggesting that independent semantic anchors better facilitate bias disentanglement. 
Finally, the ablation of key components, including the residual connection, learnable prototypes, and the multi-head attention (MHA) module, leads to substantial performance drops, confirming their necessity for robust visual bias correction.

\input{tables/table6_mod_var.tex}

\noindent\textbf{Ablation Study of R$^2$E Module.}
{To assess the effect of module ordering, we place R$^2$E either before or after the encoder while keeping VBC fixed, and also evaluate a variant with R$^2$E and VBC swapped. 
Results in Tab.~\ref{table_lsm_r2e} show that applying R$^2$E after the encoder achieves the best performance, whereas swapping it with VBC degrades accuracy. This indicates that R$^2$E is more effective when operating on well-structured representations, where it can better perform high-level reasoning and highlight critical referential cues.} 

\noindent\textbf{Ablation Study on Module Variants.} 
To further validate the superiority of our architecture, we benchmark several structural variants of VBC (Fig.~\ref{FIG_5_VBC_variant}) and R$^2$E (Fig.~\ref{FIG_5_R2E_variant}) in Tab.~\ref{table_mod_var}. 
Across all settings, the complete design achieves the highest performance (83.17\% on test) and exhibits superior accuracy, whereas altering core interactions or removing key components leads to consistent performance degradation. 
These ablation results demonstrate a consistent trend: any modification to the core interaction designs in VBC or the reasoning formulation in R$^2$E results in reduced accuracy and diminished stability, confirming that our final architecture is more effectively configured than its structural variants. Specifically, VBC var.~1/2/3 exhibit performance drops of 0.89, 1.93, and 0.45 percentage points, respectively, while R$^2$E var.~1/2/3 decrease by 0.38, 2.48, and 0.73 points. For VBC, modifying the attention interaction, substituting it with dot-product weighting, or removing semantic prototypes consistently yields sub-optimal performance. Similarly, for R$^2$E, directly reweighting language features, omitting the weighting branch, or substituting it with MHA all lead to noticeable performance regressions, thereby validating the architectural advantages of our optimized design.

\input{tables/table7_align_loss.tex}

\input{tables/table8_lora_rank.tex}

\noindent\textbf{Ablation Study of Alignment Loss.} 
Tab.~\ref{table_align_loss} evaluates the individual contributions of our alignment strategies. 
We observe that all three components positively impact grounding performance. 
Specifically, the removal of RTA or PTA results in the most significant performance drops (up to 1.89\% on val and 1.53\% on test), highlighting their critical role in capturing fine-grained spatial and regional constraints. 

\noindent\textbf{Ablation Study of LoRA Configuration.} 
Tab.~\ref{table_lora_rank} examines the impact of LoRA settings. 
Freezing all pre-trained parameters (\textit{w/o} LoRA) causes a severe performance drop (65.15\% on test, down $\sim$18 points), underscoring the necessity of task-specific adaptation. We further explore the sensitivity of the rank $r$ on the 
RefCOCOg-UMD dataset. While the performance remains relatively robust across different rank 
values, $r=32$ provides the optimal balance, achieving a peak accuracy of 83.17\% on the test 
set. 
%
These results demonstrate that BARE synergizes effectively with PEFT, adapting general pre-trained representations to referential reasoning with minimal overhead.

\begin{figure*}[htbp]\centering
	\includegraphics[scale=0.49]{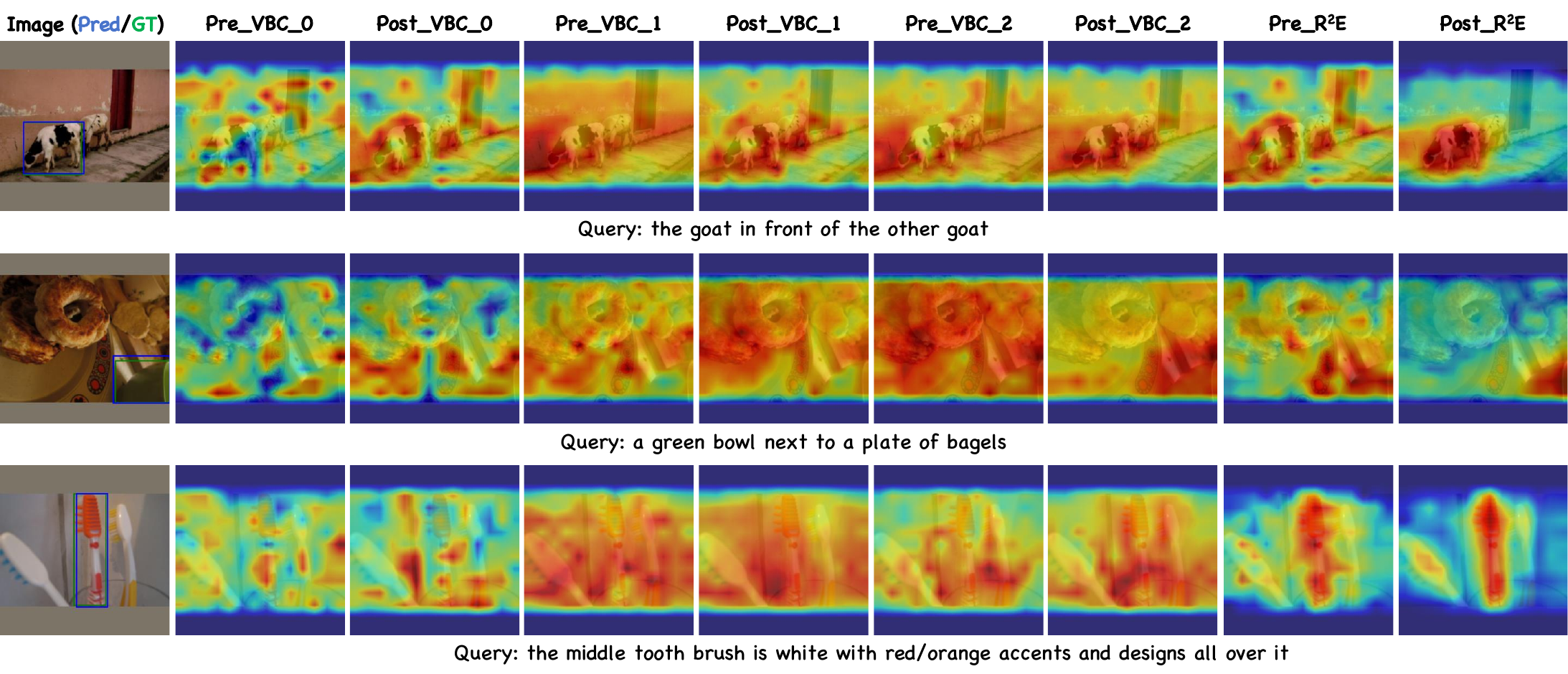}
    \vspace{-20pt}
	\caption{Layer-wise visualization of attention maps across VBC and R$^2$E modules. 
    The top row displays the original image, while subsequent rows demonstrate the attention transition: 
    VBC modules (at Layers 1, 5, and 9) progressively refocus the model's gaze from initially diffused areas onto the primary referent, 
    while the R$^2$E module (at the final encoder layer) sharply suppresses background distractions and irrelevant entities. 
    Pre and Post denote the attention states before and after the respective module processing, highlighting their effectiveness in correcting visual bias and enhancing referential reasoning.
    }
        
    \label{FIG_8}
    \vspace{-8pt}
\end{figure*}

\subsection{Further Remark}

Natural language queries in VG tasks extend well beyond category descriptions, often encoding diverse linguistic phenomena \cite{gres,ReferringExpressionCounting}, including attribute modifiers, relational constraints, etc.
Therefore, these phenomena impose distinct semantic constraints and reasoning demands, exposing different bottlenecks (\textit{e.g.}, fine-grained attribute recognition, cross-entity composition, and viewpoint alignment) and leading to substantial scenario performance variance behind aggregate metrics. 
As a result, reporting only dataset-level average accuracy cannot reveal which linguistic phenomena drive the gains or where systematic weaknesses remain. 
To this end, we build a fine-grained evaluation protocol on the RefCOCOg test set, re-partitioning samples along two axes---difficulty (easy, medium, hard) and reference types (attribute, relation, logic, ambiguity, perspective)---to quantitatively diagnose module-level robustness gains and failure modes under specific phenomena, as outlined in Template~\ref{table_prompt_tpl}.
%

\begin{figure}[t]\centering
	\includegraphics[width=8.8cm]{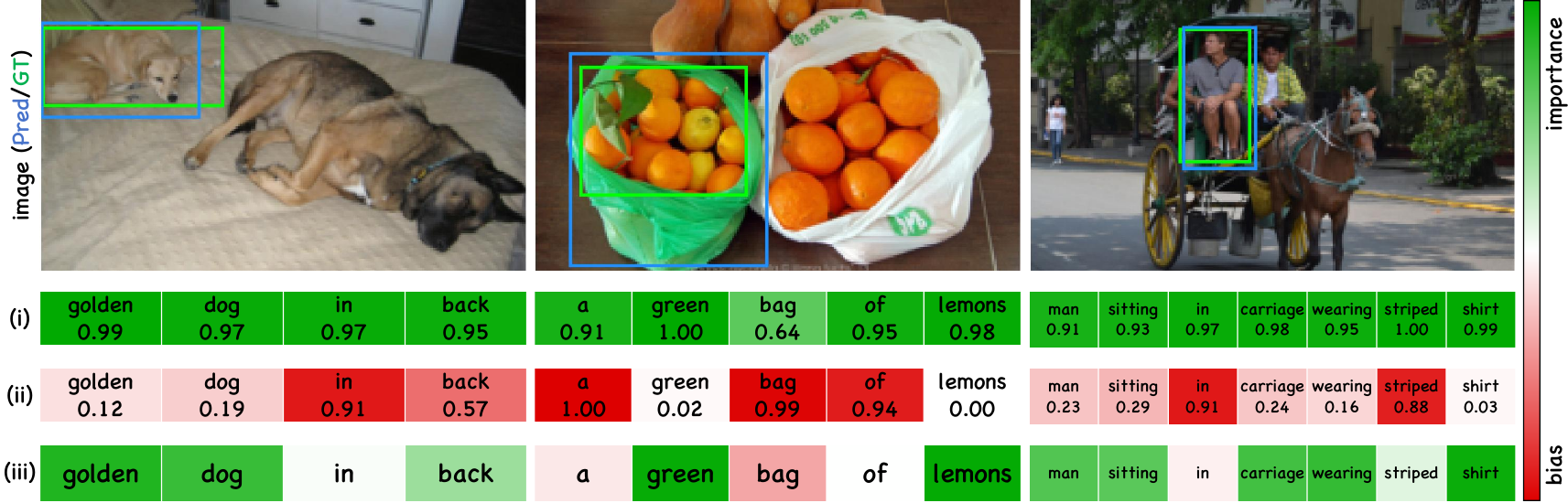}
	\caption{Visualization of the LSM mechanism across various queries. 
    The token heatmaps, arranged from top to bottom, visualize: 
    \textit{(i)} referential salience that highlights critical grounding cues; 
    \textit{(ii)} debiasing degree that identifies frequency-based linguistic shortcuts; 
    and \textit{(iii)} the final weighted result that purifies referential semantics. 
    Higher color intensity indicates a greater gating value. 
    This visualization demonstrates effectiveness of LSM in suppressing linguistic priors while accentuating task-relevant tokens.
    }
    \label{FIG_9}
    \vspace{-8pt}
\end{figure}

\noindent\textbf{Automated Partitioning Strategy.}
To enable fine-grained annotation and a reproducible partition, we use Gemini 2.5 Pro \cite{team2023gemini,team2024gemini1.5} to jointly analyze each image–query pair as follows: 
\textit{\textbf{Firstly}}, we construct a unified prompt template that explicitly defines the criteria for the five reference types and the three difficulty levels, and includes several examples to calibrate the decision boundaries.
\textit{\textbf{Secondly}}, guided by this template, Gemini parses each multimodal input and assigns a referential category and difficulty level according to the predefined rules, accompanied by a brief justification. 
\textit{\textbf{Finally}}, to facilitate downstream statistics and automated evaluation, we require the model to return the outputs in a predefined JSON schema, aligning the annotations with the original samples and ensuring that the resulting splits are structured, traceable, and reusable. 

The detailed breakdown of the 9,482 samples in the RefCOCOg test set is reported in Tab.~\ref{table_data_dist}.
Each query is assigned a single difficulty level, but it may exhibit multiple linguistic patterns. 
Accordingly, the total count remains 9,482 under the difficulty split, whereas counting by linguistic patterns yields 19,334 instances, suggesting that such phenomena are pervasive in referring queries.


\input{tables/template1.tex}


\noindent\textbf{Different Approaches Analysis.} 
We evaluate the performance of BARE against current SOTA methods, including OneRef \cite{oneref}, SimVG \cite{SimVG}, and HiVG \cite{HiVG}, across three difficulty levels and five referential categories. As summarized in Tab.~\ref{table_diff_eval}, all methods exhibit a consistent performance decline from easy to hard splits, which validates the effectiveness of our difficulty-based partitioning. Notably, BARE achieves SOTA accuracy on both the Easy (90.73\%) and Hard (71.69\%) subsets, demonstrating superior robustness and the ability to maintain high precision even as semantic complexity increases.

A more granular comparison in Fig.~\ref{FIG_6} reveals that BARE is particularly effective under challenging linguistic conditions. While BARE maintains competitive results on common attribute and relation types, it delivers pronounced improvements in logic-hard and ambiguity-easy cases. These gains can be attributed to the VBC module's capacity to suppress deceptive visual shortcuts and the R$^2$E module's strength in reinforcing high-level referential reasoning. By successfully filtering out misleading saliency cues and accentuating task-relevant tokens, BARE achieves a more stable and generalized grounding performance across diverse semantic complexities compared to its counterparts.

\noindent\textbf{Different Modules Analysis.} To quantify the standalone contribution of each module and the additional benefit from joint reasoning, Fig.~\ref{FIG_7} reports the fine-grained grounding accuracy of BARE, its ablated variants (\textit{w/o} VBC, \textit{w/o} R$^2$E, \textit{w/o} LSM), and the vanilla BEiT baseline under phenomenon- and difficulty-aware splits of \texttt{gref\_umd\_test}. 

BARE consistently outperforms four variants across all combinations, with more pronounced gains on the medium and hard splits, indicating more robust generalization as semantic complexity increases.
At the module level, removing VBC yields the most evident drops on logic-hard (down by 6.8\%), ambiguity-medium (down by 4.6\%), and perspective-hard (down by 3.2\%), indicating that visual bias correction is particularly important when dealing with distractor-heavy expressions, reference-frame changes, and logically challenging cases. In contrast, ablating R$^2$E causes broad degradation on difficult phenomena, with the largest declines on logic-hard (down by 5.9\%) and relation-hard (down by 4.5\%), which confirms that R$^2$E strengthens compositional reasoning over structured relations and logical constraints.

\input{tables/table9_data_dist}

\input{tables/table10_diff_eval}

In addition, LSM, a token-level gating mechanism applied at the input stage, generally benefits the medium or hard settings (notably on logic-medium and the hard attribute or logic splits), by filtering linguistic bias and refining semantic signals before entering the shared encoder. This reduces the interference of high-frequency patterns and co-occurrence biases on subsequent cross-modal alignment, albeit with minor variation across phenomena.

Finally, VBC and R$^2$E deliver comparable improvements across most subsets and are typically more influential than LSM when considered in isolation, while integrating all three components achieves the strongest overall performance and the most balanced gains. This trend supports strong functional complementarity: VBC corrects visual bias and purifies candidate regions, R$^2$E strengthens relational reasoning and disambiguation, and LSM mitigates language-side bias while improving cue quality. Together, they yield more comprehensive improvements and better generalization across diverse linguistic phenomena and visual conditions, further validating the effectiveness of the proposed framework.


\begin{figure*}[htbp]\centering
	\includegraphics[width=1.0\textwidth]{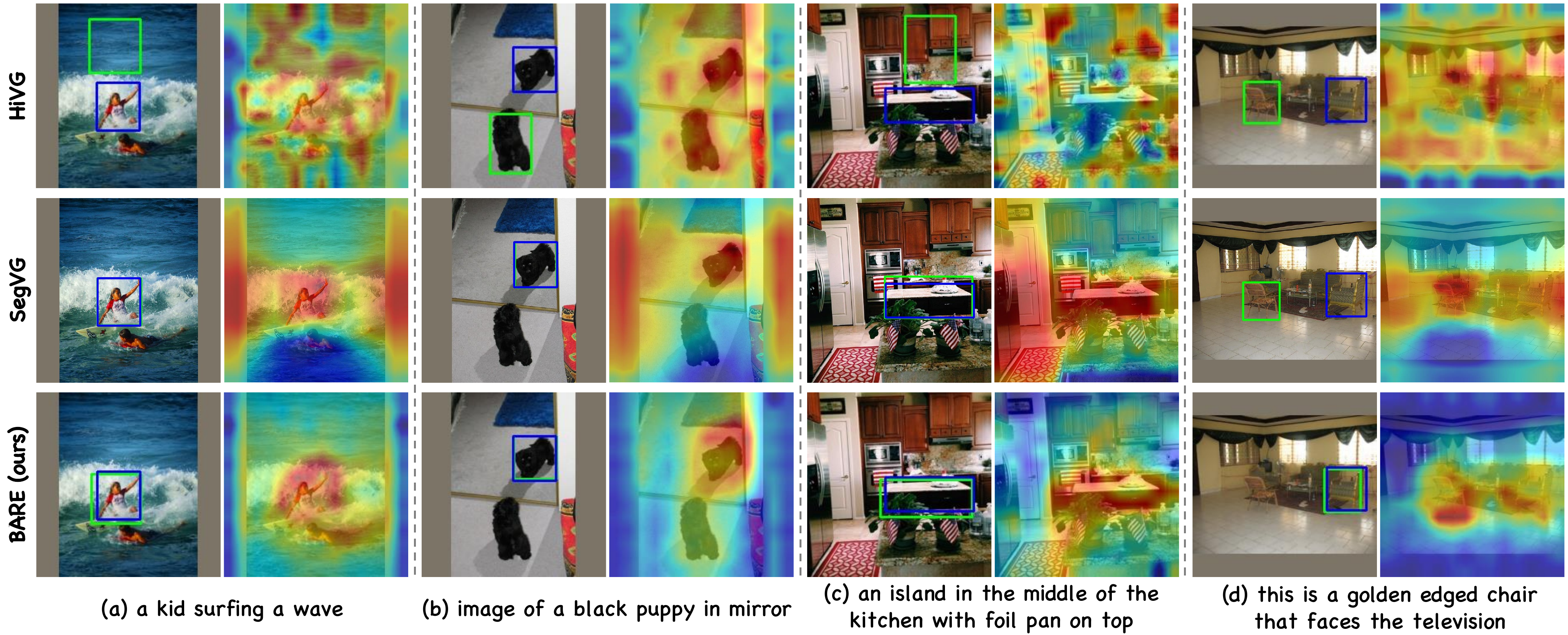}
    \vspace{-18pt}
	\caption{Qualitative comparison of HiVG, SegVG, and our BARE on RefCOCOg across four challenging scenarios. The attention maps, derived from the encoder outputs of each method, highlight BARE's robust visual perception in maintaining precise grounding within complex scenarios.
                }
        
    \label{FIG_10}
    \vspace{-8pt}
\end{figure*}

\subsection{Qualitative Results}

\noindent\textbf{Visualization of LSM.} 
We examine the internal behavior of the LSM module in Fig.~\ref{FIG_9}, where the visualization reveals a clear functional synergy through three distinct stages of linguistic processing: (\textit{i}) the salience gate identifies and highlights critical semantic cues, such as target-specific attributes and relational descriptors, that are essential for grounding. (\textit{ii}) The debiasing gate concurrently assigns weights to potential linguistic shortcuts (\textit{e.g.}, \textit{``a bag of"}) that often lead to deceptive reasoning. (\textit{iii}) The final element-wise weighted result successfully filters out redundant linguistic noise while accentuating task-relevant tokens. This progressive refinement ensures that the subsequent multimodal alignment is driven by genuine referential logic rather than linguistic co-occurrences.

\noindent\textbf{Visualization of VBC and R$^2$E.} 
To intuitively demonstrate the internal mechanism of the proposed modules, we visualize the layer-wise attention maps in Fig. \ref{FIG_8}. 
As illustrated, the VBC module effectively refocuses the model's attention from initially defocused maps onto the primary target. 
Notably, after each VBC module, the attention maps consistently exhibit high-intensity activation on the subject, which validates its effectiveness in correcting visual bias and maintaining stable target awareness across layers. 
Furthermore, the R$^2$E module, applied at the final stage, precisely eliminates irrelevant visual distractions. 
It demonstrates high target sensitivity and precision in refining referential relationships, leading to a sharp and accurate final localization. 
These visualizations confirm that BARE successfully suppresses distractions and reinforces genuine referential cues.

\noindent\textbf{Comparison with SOTA Methods.} Fig.~\ref{FIG_10} presents qualitative comparisons with HiVG and SegVG across four challenging scenarios. As illustrated, BARE consistently achieves more accurate target grounding with sharper attention focus, demonstrating robust visual perception and superior capability for fine-grained spatial and semantic reasoning.

%% file: tables/table2_comp_cost.tex
\begin{table}[t]\footnotesize
\setlength{\tabcolsep}{4pt}

\caption{Training and inference cost comparison obtained on RefCOCO dataset.  The best performance results are marked in \textbf{bold} and the second best are with \underline{underline}. FPS (frame per second) denotes the inference speed on the test set.}
\begin{center}
\begin{tabular}{c|c|c|c|c|c|c}
    \toprule
    \multirow{2}{*}{Model} & update/all  & update & Flops             &test          &testA               & testA  \\
                           &  param.     & ratio  & (G)$\downarrow$   & FPS$\uparrow$  & time$\downarrow$     & Acc.$\uparrow$  \\
    \midrule
    QRNet \cite{QRNet}           & 273/273M   & 100\%     & 250.8               & 1.98 & 1422s & 85.9\%  \\ 
    VG-LAW \cite{VG-LAW}           & 150/150M   & 100\%     & 172.8               & 3.25 & -- & 88.6\%  \\   
    CLIP-VG \cite{CLIP-VG}          & \underline{21/181M}    & \underline{12.2\%}  & \textbf{33.9}  & \textbf{14.67} &\textbf{192s}  & 87.8\%  \\  
    HiVG-B \cite{HiVG}           & 41/206M    & 20.1\%    & \underline{38.7}    & \underline{13.77}     &\underline{205s} & \underline{89.9\% } \\  
    \midrule
    \rowcolor{gray!10}
    BARE-B          & \textbf{4.7/276M}& \textbf{1.7\%}  & 78.5    &11.08 & 255s  & \textbf{91.6\% } \\
    \bottomrule
\end{tabular}%

\end{center}
\label{table_comp_cost}
\vspace{-8pt}
\end{table}

%% file: tables/table3_main_ablation.tex
\begin{table}[t]\footnotesize
\setlength{\tabcolsep}{10pt}
\centering

\caption{Ablation study on RefCOCOg-UMD evaluating the individual contributions of Linguistic Salience Modulator (LSM), Visual Bias Correction (VBC), Referential Relationship Enhancement (R$^2$E), and the Fusion Encoder.}
\label{table_main_ablation}
\begin{tabular}{c|c|c|c c}
\toprule
\multirow{2}{*}{LSM} &\multirow{2}{*}{VBC} &\multirow{2}{*}{R$^2$E} &\multicolumn{2}{c}{RefCOCOg(Accu@0.5)} \\ 
                     &                        &                                &\textit{val}        &\textit{test} \\
\midrule

\ding{52}    &  \ding{56}    &\ding{56}      & 81.47   & 81.79  \\
\ding{56}    &  \ding{52}    &\ding{56}      & 81.84   & 82.42 \\
\ding{56}    &  \ding{56}    &\ding{52}      & 81.70   & 82.04 \\
\rowcolor{gray!10}
\ding{52}    &  \ding{52}    &\ding{52}      & 82.73   & 83.17 \\ 
\midrule
\multicolumn{3}{c|}{\textit{w/o} Fusion Encoder}     & 79.02   & 79.38 \\

\bottomrule
\end{tabular}%
\end{table}

%% file: tables/table4_vbc_ablation.tex








\begin{table}[t]
\centering
\setlength{\tabcolsep}{8pt}
\caption{{Ablation results of VBC on RefCOCOg-umd. We evaluate the effects of stride size, insertion depth, prototype initialization (derived from visual features or randomly), and key module components. The term \textit{w/o} denotes without.}}
\label{table_vbc_ablation}
\begin{tabular}{c|c|c|c}
\toprule
Stride size & Insertion depth & \textit{val} & \textit{test} \\
\midrule
2 & Layer1 $\sim$ Layer2 & 82.05 & 82.60 \\
3 & Layer1 $\sim$ Layer2 & 81.96 & 82.21 \\
\rowcolor{gray!10}
4 & Layer1 $\sim$ Layer2 & 82.73 & 83.17 \\
\midrule
\rowcolor{gray!10}
4 & Layer1 $\sim$ Layer2 & 82.73 & 83.17 \\
4 & Layer2 $\sim$ Layer3 & 82.33 & 82.53 \\
4 & Layer3 $\sim$ Layer4 & 82.18 & 82.47 \\
\midrule
\midrule
\multicolumn{2}{c|}{Mean-pooling initialization} & 82.37 & 82.69 \\
\multicolumn{2}{c|}{Max-pooling initialization} & 82.51 & 82.90 \\
\rowcolor{gray!10}
\multicolumn{2}{c|}{Random initialization} & 82.73 & 83.17 \\
\midrule
\multicolumn{2}{c|}{VBC \textit{w/o} residual input} & 81.56 & 80.96 \\
\multicolumn{2}{c|}{VBC \textit{w/o} learnable prototype} & 81.63 & 81.76 \\
\multicolumn{2}{c|}{VBC \textit{w/o} MHA module} & 81.31 & 81.87 \\
\bottomrule
\end{tabular}
\end{table}

%% file: tables/table5_lsm_r2e.tex
\begin{table}[t]
\setlength{\tabcolsep}{8pt}
\centering
\caption{Ablation study on RefCOCOg-UMD: (Top) effect of the dual-gate mechanism in LSM, where \textit{w/o} denotes removing a specific gate; (Bottom) effect of R$^2$E insertion position relative to the encoder.}
\label{table_lsm_r2e}
\begin{tabular}{l|cc}
    \toprule
    LSM Modulating Strategy & \textit{val} & \textit{test} \\
    \midrule
    \textit{w/o} salience gate       & 82.27   & 82.50  \\
    \textit{w/o} debiasing gate                  & 82.16   & 82.22  \\
    \rowcolor{gray!10}
    \textit{w/ }  both gate       & 82.73   & 83.17  \\
    \midrule
    \midrule
R$^2$E Insertion Strategy & \textit{val} & \textit{test} \\
\midrule
Inserted before the encoder (first layer)       & 82.05   & 82.60  \\
Swapped with VBC   (intermediate layer)         & 81.96   & 82.21  \\
\rowcolor{gray!10}
Inserted after the encoder (last layer)         & 82.73   & 83.17  \\     
\bottomrule
\end{tabular}
\vspace{-8pt}
\end{table}

%% file: tables/table6_mod_var.tex
\begin{table}[t]\footnotesize
\setlength{\tabcolsep}{12pt}
\centering
\caption{Ablation study on architectural variants of VBC and R$^2$E modules on the RefCOCOg-UMD dataset, where \textit{var.} denotes structural variations of the proposed modules.}
\label{table_mod_var}
\begin{tabular}{l|l|c|c}
\toprule
\textbf{VBC Module} & \textbf{R$^2$E Module} & \textit{val} & \textit{test} \\
\midrule
VBC var. 1 & R$^2$E & 82.29 & 82.28 \\
VBC var. 2 & R$^2$E & 80.96 & 81.24 \\
VBC var. 3 & R$^2$E & 82.14 & 82.72 \\
VBC & R$^2$E var. 1 & 82.37 & 82.79 \\
VBC & R$^2$E var. 2 & 80.62 & 80.69 \\
VBC & R$^2$E var. 3 & 82.10 & 82.44 \\
\rowcolor{gray!10}
VBC & R$^2$E & 82.73 & 83.17 \\
\bottomrule
\end{tabular}
\vspace{-8pt}    
\end{table}

%% file: tables/table7_align_loss.tex
\begin{table}[t]
\setlength{\tabcolsep}{8pt}
\centering
\caption{Ablation study of multi-granularity alignment strategies. ITA, RTA, and PTA represent image-text, region-text, and pixel-text alignment, respectively. Parentheses indicate the performance drop relative to the full model.}
\label{table_align_loss}
\begin{tabular}{c | c | c|c c}
\toprule

\textbf{ITA} & \textbf{RTA} & \textbf{PTA} & \textit{val} & \textit{test} \\
\midrule

{\ding{56}}    & {\ding{52}}     & {\ding{52}}   & 81.55 ($\downarrow$~1.18) & 81.92 ($\downarrow$~1.25)  \\

{\ding{52}}    & {\ding{56}}     & {\ding{52}}   & 80.84 ($\downarrow$~1.89) & 81.72 ($\downarrow$~1.45) \\

{\ding{52}}    & {\ding{52}}     & {\ding{56}}   & 81.19 ($\downarrow$~1.54) & 81.64 ($\downarrow$~1.53) \\
\rowcolor{gray!10}
{\ding{52}}    & {\ding{52}}     & {\ding{52}}   & {82.73}   & {83.17}  \\

\bottomrule
\end{tabular}%
\vspace{-8pt}

\end{table}

%% file: tables/table8_lora_rank.tex
\begin{table}[t]\footnotesize
\renewcommand{\arraystretch}{1}
\setlength{\tabcolsep}{8pt}
\centering
\caption{Ablation study on LoRA configurations on RefCOCOg-UMD. $r$ and $\alpha$ denote the rank and scaling factor, respectively. The gray-shaded row indicates our final configuration.}

\label{table_lora_rank}

\begin{tabular}{c|c c}
\toprule

Architecture                       &val        &test \\
\midrule

BARE \textit{w/o} LoRA (all params frozen) & 64.54   & 65.15  \\
BARE \textit{w/} LoRA ($\alpha$=16, r=16)          & 82.19   & 82.07  \\
\rowcolor{gray!10}
BARE \textit{w/} LoRA ($\alpha$=16, r=32)          & 82.73   & 83.17  \\
BARE \textit{w/} LoRA ($\alpha$=16, r=64)          & 82.49   & 82.68  \\ 

\bottomrule
\end{tabular}%

\vspace{-8pt}
\end{table}

%% file: tables/template1.tex
\newcounter{template}
\renewcommand{\thetemplate}{\Roman{template}}
\begin{table}[h]
    \refstepcounter{template}
    \label{table_prompt_tpl}
\begin{tcolorbox}[
    colback=gray!5,
    colframe=black!70,
    title=\textbf{Template~\thetemplate. Prompt for Automated Partitioning.},
    fonttitle=\bfseries,
    boxrule=0.8pt,
    arc=2mm
  ]
  
  \textbf{Role.}
  You are an expert AI assistant for referring expression understanding and difficulty analysis.
  
  \textbf{Input.}
  Each instance consists of:
  \begin{itemize}[leftmargin=1.5em]
    \item an image
    \item a target bounding box formatted as $\,[x, y, w, h]\,$
    \item a natural language expression describing the target
  \end{itemize}
  
  \textbf{Referential Type.}
  You must analyze the phrase and judge whether it involves the following five types
of linguistic phenomena that require non-trivial reasoning for precise grounding:
    \begin{enumerate}[leftmargin=1.8em]
        \item \textbf{Attribute}: Focuses on intrinsic visual traits such as color, size, material, or state (\textit{e.g.}, ``the small red ball'', ``the wooden table''). It captures fundamental properties of the target itself.
        \item \textbf{Relation}: Localizes the target via its extrinsic links to landmarks, including spatial relations (\textit{e.g.}, ``next to'', ``above'') and interactive actions (\textit{e.g.}, ``holding a bag'', ``sitting on'').
        \item \textbf{Logic}: Involves logical operations to select candidates, such as negation (\textit{e.g.}, ``without a collar''), conjunction (\textit{e.g.}, ``in red and wearing glasses''), or exclusivity (\textit{e.g.}, ``all others except'').
        \item \textbf{Ambiguity}: Demands nuanced discrimination to resolve referential uncertainty in cluttered scenes (\textit{e.g.}, ``the person in a white shirt'' in a crowd), or when interpreting subjective and abstract descriptors (\textit{e.g.}, ``crazy colors'', ``blurry pizza'').
        \item \textbf{Perspective}: Requires 3D physical spatial reasoning to interpret relative viewpoints (\textit{e.g.}, ``facing the camera'') or mirrored configurations (\textit{e.g.}, ``the puppy in the mirror''), where coordinates transcend simple 2D image axes.
  \end{enumerate}
  
  \textbf{Output.}
  A structured JSON object indicating:
  \begin{itemize}[leftmargin=1.5em]
    \item the presence of each reasoning type
    \item an overall difficulty level (\textit{easy}, \textit{medium}, \textit{hard})
    \item a concise rationale for the reasoning and difficulty assessments
  \end{itemize}
  
  \textbf{Judgment Principles.}
  \textit{(i)} Final assessments should be grounded in a holistic interpretation of the interplay between the referring expression and the visual context. 
  \textit{(ii)} The type is identified as present only if it is essential for the precise disambiguation of the target.
  
  \end{tcolorbox}
  \vspace{-12pt}
\end{table}

%% file: tables/table9_data_dist.tex
\begin{table}[t]
\centering
\caption{Statistics of the fine-grained evaluation splits on RefCOCOg-UMD test set. Each sample is assigned one difficulty level, while referential categories may overlap across samples.}
\begin{tabular}{c|l|c|c|c}
\toprule
\textbf{Category} & \textbf{Subset} & \textbf{Avg. Length} & \textbf{Samples} & \textbf{Total} \\
\midrule
\multirow{3}{*}{Difficulty} 
    & Easy      & 33.85 & 4,109 & \multirow{3}{*}{9,482} \\
    & Medium    & 46.19 & 4,517 &  \\
    & Hard      & 47.87 & 856   &  \\
\midrule
\multirow{5}{*}{\shortstack{Referential \\ Type}} 
    & Attribute              & 43.33 & 7,070 & \multirow{5}{*}{19,334} \\
    & Relation               & 44.48 & 7,474 &  \\
    & Logic             & 53.89 & 666   &  \\
    & Ambiguity              & 42.84 & 2,616 &  \\
    & Perspective       & 42.79 & 1,508 &  \\

\bottomrule
\end{tabular}
\label{table_data_dist}
\vspace{-8pt}
\end{table}

%% file: tables/table10_diff_eval.tex
\begin{table}[t]
\centering
\setlength{\tabcolsep}{8pt}
\caption{Performance comparison across difficulty levels on RefCOCOg test set. Accuracy (\%) is reported for each method under easy, medium, and hard splits. $^{\dagger}$ denotes results reproduced by us under the same experimental settings.}
\begin{tabular}{c|l|c|c|c|c}
\toprule
\textbf{Category} & \textbf{Subset} & \textbf{HiVG} & \textbf{SimVG$^{\dagger}$} & \textbf{OneRef} & \textbf{BARE} \\
\midrule
\multirow{3}{*}{Difficulty} 
    & Easy      & 86.96 & 85.81 & 90.56 & \textbf{90.73} \\
    & Medium    & 71.73 & 68.72 & \textbf{79.36} & 78.42 \\
    & Hard      & 63.20 & 57.00 & {71.63} & \textbf{71.69} \\
\bottomrule
\end{tabular}
\label{table_diff_eval}
\vspace{-8pt}
\end{table}

%% file: Body/5-futurework.tex
In the future, as a bias-aware and reasoning-enhanced paradigm for one-tower visual grounding, BARE holds strong potential for broader applications across diverse downstream tasks. Beyond standard grounding benchmarks, it can be extended to more semantically complex scenarios, such as General Referring Expression Comprehension (GREC) \cite{gres}, which involves diverse and ambiguous referential expressions, and Referring Image Editing (RIE) \cite{ReferringImageEditing, smartedit}, which requires fine-grained localization to guide pixel-level manipulations. 
These directions offer promising avenues to further exploit BARE’s capabilities in referential reasoning and bias mitigation for real-world multimodal applications.

%% file: Body/6-conclusion.tex
To address the limitations of over-entangled features and insufficient semantic reasoning in complex grounding tasks, we propose BARE, a bias-aware and reasoning-enhanced framework for one-tower visual grounding.
BARE introduces the language salience modulator, visual bias correction and referential relationship enhancement modules to simultaneously suppress deceptive multimodal biases and strengthen referential understanding, enabling  more precise grounding performance by filtering out multimodal distractions and focusing on critical cues.
Extensive experiments on five benchmarks demonstrate the effectiveness of BARE, while ablation studies confirm the necessity of each component. 
Future work will explore its extension to multi-image and general grounding.

%% file: references.bib
@article{VisualGroundingWithDualKnowledgeDistillation,
  title     = {Visual grounding with dual knowledge distillation},
  author    = {Wu, Wansen and Cao, Meng and Hu, Yue and others},
  journal   = {IEEE Trans. Circuits Syst. Video Technol.},
  volume    = {34},
  number    = {10},
  pages     = {10399--10410},
  year      = {2024},
  publisher = {IEEE}
}

@article{One-peace,
  title   = {One-peace: Exploring one general representation model toward unlimited modalities},
  author  = {Wang, Peng and Wang, Shijie and Lin, Junyang and others},
  journal = {arXiv preprint arXiv:2305.11172},
  year    = {2023}
}

@article{MFSD,
  title     = {A Masked Reference Token Supervision based Iterative Visual-language Framework for Robust Visual Grounding},
  author    = {Wang, Chunlei and Feng, Wenquan and Lyu, Shuchang and others},
  journal   = {IEEE Trans. Circuits Syst. Video Technol.},
  year      = {2024},
  publisher = {IEEE}
}

@inproceedings{GVQA,
  title     = {Don't just assume; look and answer: Overcoming priors for visual question answering},
  author    = {Agrawal, Aishwarya and Batra, Dhruv and Parikh, Devi and Kembhavi, Aniruddha},
  booktitle = {Proc. IEEE/CVF Conf. Comput. Vis. Pattern Recognit. (CVPR)},
  pages     = {4971--4980},
  year      = {2018}
}

@article{RUBi,
  title   = {Rubi: Reducing unimodal biases for visual question answering},
  author  = {Cadene, Remi and Dancette, Corentin and Cord, Matthieu and Parikh, Devi},
  journal = {in Proc. Adv. Neural Inf. Process. Syst. (NeurIPS)},
  volume  = {32},
  year    = {2019}
}

@inproceedings{DeepFeatureInterpolation,
  title     = {Deep feature interpolation for image content changes},
  author    = {Upchurch, Paul and Gardner, Jacob and Pleiss, Geoff and others},
  booktitle = {Proc. IEEE/CVF Conf. Comput. Vis. Pattern Recognit. (CVPR)},
  pages     = {7064--7073},
  year      = {2017}
}

@inproceedings{UnveilingInternalReasoningModes,
  title     = {Unveiling Internal Reasoning Modes in LLMs: A Deep Dive into Latent Reasoning vs. Factual Shortcuts with Attribute Rate Ratio},
  author    = {Yang, Yiran and Sun, Haifeng and Wang, Jingyu and others},
  booktitle = {Proc. Conf. Empirical Methods Natural Lang. Process. (EMNLP)},
  pages     = {2186--2206},
  year      = {2025}
}

@inproceedings{ModalityCompetition,
  title        = {Modality competition: What makes joint training of multi-modal network fail in deep learning},
  author       = {Huang, Yu and others},
  booktitle    = {Proc. Int. Conf. Mach. Learn. (ICML)},
  pages        = {9226--9259},
  year         = {2022},
  organization = {PMLR}
}

@article{QuestionConditionedDebiasing,
  title     = {Question-conditioned debiasing with focal visual context fusion for visual question answering},
  author    = {Liu, Jin and Wang, Guoxiang and Fan, Chongfeng and Zhou, Fengyu and Xu, Huijuan},
  journal   = {Knowledge-Based Syst.},
  volume    = {278},
  pages     = {110879},
  year      = {2023},
  publisher = {Elsevier}
}

@inproceedings{AutomaticallyNeutralizingSubjectiveBiasInText,
  title     = {Automatically neutralizing subjective bias in text},
  author    = {Pryzant, Reid and Martinez, Richard Diehl and Dass, Nathan and others},
  booktitle = {Proc. AAAI Conf. Artif. Intell.},
  volume    = {34},
  number    = {01},
  pages     = {480--489},
  year      = {2020}
}

@article{Coop,
  title     = {Understanding and mitigating overfitting in prompt tuning for vision-language models},
  author    = {Ma, Chengcheng and Liu, Yang and Deng, Jiankang and others},
  journal   = {IEEE Trans. Circuits Syst. Video Technol.},
  volume    = {33},
  number    = {9},
  pages     = {4616--4629},
  year      = {2023},
  publisher = {IEEE}
}

@article{PLVL,
  title   = {Progressive Language-guided Visual Learning for Multi-Task Visual Grounding},
  author  = {Wang, Jingchao and Wang, Hong and Zhang, Wenlong and others},
  journal = {arXiv preprint arXiv:2504.16145},
  year    = {2025}
}

@article{PAR,
  title     = {Pedestrian attribute recognition via clip based prompt vision-language fusion},
  author    = {Wang, Xiao and Jin, Jiandong and Li, Chenglong and others},
  journal   = {IEEE Trans. Circuits Syst. Video Technol.},
  year      = {2024},
  publisher = {IEEE}
}

@article{ZS-TAD,
  title     = {Zero-shot temporal action detection by learning multimodal prompts and text-enhanced actionness},
  author    = {Raza, Asif and others},
  journal   = {IEEE Trans. Circuits Syst. Video Technol.},
  volume    = {34},
  number    = {11},
  pages     = {11000--11012},
  year      = {2024},
  publisher = {IEEE}
}

@inproceedings{TTE,
  title     = {TTE: Two Tokens are Enough to Improve Parameter-Efficient Tuning},
  author    = {Ruan, Jiacheng and Xie, Mingye and Gao, Jingsheng and others},
  booktitle = {Proc. AAAI Conf. Artif. Intell.},
  volume    = {39},
  number    = {19},
  pages     = {20209--20217},
  year      = {2025}
}

@article{PBAL,
  title     = {Prompt-augmented Boundary Attentive Learning for Weakly Supervised Temporal Sentence Grounding},
  author    = {Zhu, Zhehao and Huang, Yifei and Zhang, Mingfang and Ouyang, Liangyang and Sato, Yoichi},
  journal   = {IEEE Trans. Circuits Syst. Video Technol.},
  year      = {2025},
  publisher = {IEEE}
}

@inproceedings{ReferringExpressionCounting,
  title     = {Referring expression counting},
  author    = {Dai, Siyang and Liu, Jun and Cheung, Ngai-Man},
  booktitle = {Proc. IEEE/CVF Conf. Comput. Vis. Pattern Recognit. (CVPR)},
  pages     = {16985--16995},
  year      = {2024}
}

@inproceedings{gres,
  title     = {Gres: Generalized referring expression segmentation},
  author    = {Liu, Chang and Ding, Henghui and Jiang, Xudong},
  booktitle = {Proc. IEEE/CVF Conf. Comput. Vis. Pattern Recognit. (CVPR)},
  pages     = {23592--23601},
  year      = {2023}
}

@inproceedings{ReinforcedSequenceTrainingBasedSubjectiveBiasCorrection,
  title     = {Reinforced sequence training based subjective bias correction},
  author    = {Madanagopal, Karthic and Caverlee, James},
  booktitle = {Proc. Conf. Eur. Chapter Assoc. Comput. Linguist. (EACL)},
  pages     = {2585--2598},
  year      = {2023}
}

@article{InstanceVG,
  title     = {Improving generalized visual grounding with instance-aware joint learning},
  author    = {Dai, Ming and others},
  journal   = {IEEE Trans. Pattern Anal. Mach. Intell.},
  year      = {2025},
  publisher = {IEEE}
}

@inproceedings{ResNet,
  title     = {Deep residual learning for image recognition},
  author    = {He, Kaiming and Zhang, Xiangyu and Ren, Shaoqing and Sun, Jian},
  booktitle = {Proc. IEEE/CVF Conf. Comput. Vis. Pattern Recognit. (CVPR)},
  pages     = {770--778},
  year      = {2016}
}

@inproceedings{SwinTransformer,
  title     = {Swin transformer: Hierarchical vision transformer using shifted windows},
  author    = {Liu, Ze and Lin, Yutong and Cao, Yue and others},
  booktitle = {Proc. IEEE/CVF Int. Conf. Comput. Vis. (ICCV)},
  pages     = {10012--10022},
  year      = {2021}
}

@inproceedings{COCO,
  title     = {Microsoft coco: Common objects in context},
  author    = {Lin, Tsung-Yi and Maire, Michael and Belongie, Serge and others},
  booktitle = {Proc. Eur. Conf. Comput. Vis. (ECCV)},
  pages     = {740--755},
  year      = {2014}
}

@article{USRI-REID,
  title     = {Adaptive Pseudo-label Purification and Debiasing for Unsupervised Visible-Infrared Person Re-Identification},
  author    = {Yin, Xiangbo and Shi, Jiangming and Zhang, Zhizhong and Xie, Yuan and Qu, Yanyun},
  journal   = {IEEE Trans. Circuits Syst. Video Technol.},
  year      = {2025},
  publisher = {IEEE}
}

@article{GBT,
  title     = {Overcoming Modality Bias in Question-Driven Sign Language Video Translation},
  author    = {Gao, Liqing and Lyu, Fan and Shi, Peng and others},
  journal   = {IEEE Trans. Circuits Syst. Video Technol.},
  volume    = {34},
  number    = {11},
  pages     = {11724--11738},
  year      = {2024},
  publisher = {IEEE}
}

@article{M2IST,
  title     = {M2ist: Multi-modal interactive side-tuning for efficient referring expression comprehension},
  author    = {Liu, Xuyang and Liu, Ting and Huang, Siteng and others},
  journal   = {IEEE Trans. Circuits Syst. Video Technol.},
  year      = {2025},
  publisher = {IEEE}
}

@inproceedings{vqa,
  title     = {Vqa: Visual question answering},
  author    = {Antol, Stanislaw and Agrawal, Aishwarya and Lu, Jiasen and others},
  booktitle = {Proc. IEEE Int. Conf. Comput. Vis. (ICCV)},
  pages     = {2425--2433},
  year      = {2015}
}

@inproceedings{smartedit,
  title     = {Smartedit: Exploring complex instruction-based image editing with multimodal large language models},
  author    = {Huang, Yuzhou and Xie, Liangbin and Wang, Xintao and others},
  booktitle = {Proc. IEEE/CVF Conf. Comput. Vis. Pattern Recognit. (CVPR)},
  pages     = {8362--8371},
  year      = {2024}
}

@inproceedings{ReferringImageEditing,
  title     = {Referring image editing: Object-level image editing via referring expressions},
  author    = {Liu, Chang and Li, Xiangtai and Ding, Henghui},
  booktitle = {Proc. IEEE/CVF Conf. Comput. Vis. Pattern Recognit. (CVPR)},
  pages     = {13128--13138},
  year      = {2024}
}

@inproceedings{Refcoco,
  title        = {Modeling context in referring expressions},
  author       = {Yu, Licheng and Poirson, Patrick and others},
  booktitle    = {Proc. Eur. Conf. Comput. Vis. (ECCV)},
  pages        = {69--85},
  year         = {2016}
}

@inproceedings{BEiT,
  title     = {Image as a foreign language: Beit pretraining for vision and vision-language tasks},
  author    = {Wang, Wenhui and Bao, Hangbo and Dong, Li and others},
  booktitle = {Proc. IEEE/CVF Conf. Comput. Vis. Pattern Recognit. (CVPR)},
  pages     = {19175--19186},
  year      = {2023}
}

@inproceedings{Refcocog,
  title     = {Generation and comprehension of unambiguous object descriptions},
  author    = {Mao, Junhua and Huang, Jonathan and Toshev, Alexander and others},
  booktitle = {Proc. IEEE/CVF Conf. Comput. Vis. Pattern Recognit. (CVPR)},
  pages     = {11--20},
  year      = {2016}
}

@inproceedings{Refcocog-umd,
  title        = {Modeling context between objects for referring expression understanding},
  author       = {Nagaraja, Varun K and Morariu, Vlad I and Davis, Larry S},
  booktitle    = {Proc. Eur. Conf. Comput. Vis. (ECCV)},
  pages        = {792--807},
  year         = {2016}
}

@inproceedings{ReferIt,
  title     = {Referitgame: Referring to objects in photographs of natural scenes},
  author    = {Kazemzadeh, Sahar and Ordonez, Vicente and others},
  booktitle = {Proc. Conf. Empirical Methods Natural Lang. Process. (EMNLP)},
  pages     = {787--798},
  year      = {2014}
}

@article{oneref,
  title={Oneref: Unified one-tower expression grounding and segmentation with mask referring modeling},
  author={Xiao, Linhui and Yang, Xiaoshan and Peng, Fang and Wang, Yaowei and Xu, Changsheng},
  journal={in Proc. Adv. Neural Inf. Process. Syst. (NeurIPS)},
  volume={37},
  pages={139854--139885},
  year={2024}
}

@article{vg_survey,
  title   = {Towards Visual Grounding: A Survey},
  author  = {Xiao, Linhui and Yang, Xiaoshan and Lan, Xiangyuan and Wang, Yaowei and Xu, Changsheng},
  journal = {IEEE Trans. Pattern Anal. Mach. Intell.},
  year    = {2025}
}

@inproceedings{flickr30k,
  title     = {Flickr30k entities: Collecting region-to-phrase correspondences for richer image-to-sentence models},
  author    = {Plummer, Bryan A and others},
  booktitle = {Proc. IEEE Int. Conf. Comput. Vis. (ICCV)},
  pages     = {2641--2649},
  year      = {2015}
}

@inproceedings{generalVQA,
  title     = {On the general value of evidence, and bilingual scene-text visual question answering},
  author    = {Wang, Xinyu and Liu, Yuliang and Shen, Chunhua and others},
  booktitle = {Proc. IEEE/CVF Conf. Comput. Vis. Pattern Recognit. (CVPR)},
  pages     = {10126--10135},
  year      = {2020}
}

@article{VD,
  title   = {Multimodal incremental transformer with visual grounding for visual dialogue generation},
  author  = {Chen, Feilong and Meng, Fandong and Chen, Xiuyi and Li, Peng and Zhou, Jie},
  journal = {arXiv preprint arXiv:2109.08478},
  year    = {2021}
}

@inproceedings{BehindTheScene,
  title        = {Behind the scene: Revealing the secrets of pre-trained vision-and-language models},
  author       = {Cao, Jize and Gan, Zhe and Cheng, Yu and others},
  booktitle    = {Proc. Eur. Conf. Comput. Vis. (ECCV)},
  pages        = {565--580},
  year         = {2020}
}

@article{AnalyzingTRM,
  title   = {Analyzing transformers in embedding space},
  author  = {Dar, Guy and Geva, Mor and Gupta, Ankit and Berant, Jonathan},
  journal = {arXiv preprint arXiv:2209.02535},
  year    = {2022}
}

@article{challenge1,
  title   = {A Closer Look at Multimodal Representation Collapse},
  author  = {Chaudhuri, Abhra and Dutta, Anjan and Bui, Tu and others},
  journal = {arXiv preprint arXiv:2505.22483},
  year    = {2025}
}

@article{MindTheGap,
  title   = {Mind the gap: Understanding the modality gap in multi-modal contrastive representation learning},
  author  = {Liang, Victor Weixin and Zhang, Yuhui and Kwon, Yongchan and Yeung, Serena and Zou, James Y},
  journal = {in Proc. Adv. Neural Inf. Process. Syst. (NeurIPS)},
  volume  = {35},
  pages   = {17612--17625},
  year    = {2022}
}

@inproceedings{challenge2,
  title     = {Multi-attribute interactions matter for 3d visual grounding},
  author    = {Xu, Can and Han, Yuehui and Xu, Rui and others},
  booktitle = {Proc. IEEE/CVF Conf. Comput. Vis. Pattern Recognit. (CVPR)},
  pages     = {17253--17262},
  year      = {2024}
}

@article{attention_is_all_you_need,
  title   = {Attention is all you need},
  author  = {Vaswani, Ashish and Shazeer, Noam and Parmar, Niki and others},
  journal = {in Proc. Adv. Neural Inf. Process. Syst. (NeurIPS)},
  volume  = {30},
  year    = {2017}
}

@inproceedings{fast-rcnn,
  title     = {Fast r-cnn},
  author    = {Girshick, Ross},
  booktitle = {Proc. IEEE Int. Conf. Comput. Vis. (ICCV)},
  pages     = {1440--1448},
  year      = {2015}
}

@inproceedings{giou,
  title     = {Generalized intersection over union: A metric and a loss for bounding box regression},
  author    = {Rezatofighi, Hamid and Tsoi, Nathan and others},
  booktitle = {Proc. IEEE/CVF Conf. Comput. Vis. Pattern Recognit. (CVPR)},
  pages     = {658--666},
  year      = {2019}
}

@inproceedings{focal,
  title     = {Focal loss for dense object detection},
  author    = {Lin, Tsung-Yi and Goyal, Priya and Girshick, Ross and He, Kaiming and Doll{\'a}r, Piotr},
  booktitle = {Proc. IEEE Int. Conf. Comput. Vis. (ICCV)},
  pages     = {2980--2988},
  year      = {2017}
}

@inproceedings{dice,
  title        = {V-net: Fully convolutional neural networks for volumetric medical image segmentation},
  author       = {Milletari, Fausto and Navab, Nassir and Ahmadi, Seyed-Ahmad},
  booktitle    = {Proc. Int. Conf. 3D Vis. (3DV)},
  pages        = {565--571},
  year         = {2016},
  organization = {IEEE}
}

@inproceedings{TransVG,
  title     = {Transvg: End-to-end visual grounding with transformers},
  author    = {Deng, Jiajun and Yang, Zhengyuan and Chen, Tianlang and Zhou, Wengang and Li, Houqiang},
  booktitle = {Proc. IEEE/CVF Int. Conf. Comput. Vis. (ICCV)},
  pages     = {1769--1779},
  year      = {2021}
}

@article{bart,
  title   = {Bart: Denoising sequence-to-sequence pre-training for natural language generation, translation, and comprehension},
  author  = {Lewis, Mike and Liu, Yinhan and Goyal, Naman and others},
  journal = {arXiv preprint arXiv:1910.13461},
  year    = {2019}
}

@article{sentencepiece,
  title   = {Sentencepiece: A simple and language independent subword tokenizer and detokenizer for neural text processing},
  author  = {Kudo, Taku and Richardson, John},
  journal = {arXiv preprint arXiv:1808.06226},
  year    = {2018}
}

@inproceedings{Agrawal2018,
  title     = {Don't just assume; look and answer: Overcoming priors in visual question answering},
  author    = {Agrawal, Aishwarya and others},
  booktitle = {Proc. IEEE/CVF Conf. Comput. Vis. Pattern Recognit. (CVPR)},
  pages     = {4971--4980},
  year      = {2018}
}

@inproceedings{CSS,
  title     = {Counterfactual samples synthesizing for robust visual question answering},
  author    = {Chen, Long and Yan, Xin and Xiao, Jun and others},
  booktitle = {Proc. IEEE/CVF Conf. Comput. Vis. Pattern Recognit. (CVPR)},
  pages     = {10800--10809},
  year      = {2020}
}

@inproceedings{YORO,
  title        = {Yoro-lightweight end to end visual grounding},
  author       = {Ho, Chih-Hui and Appalaraju, Srikar and Jasani, Bhavan and Manmatha, R and Vasconcelos, Nuno},
  booktitle    = {Proc. Eur. Conf. Comput. Vis. (ECCV)},
  pages        = {3--23},
  year         = {2022}
}

@inproceedings{VLTVG,
  title     = {Improving visual grounding with visual-linguistic verification and iterative reasoning},
  author    = {Yang, Li and Xu, Yan and Yuan, Chunfeng and others},
  booktitle = {Proc. IEEE/CVF Conf. Comput. Vis. Pattern Recognit. (CVPR)},
  pages     = {9499--9508},
  year      = {2022}
}

@inproceedings{QRNet,
  title     = {Shifting more attention to visual backbone: Query-modulated refinement networks for end-to-end visual grounding},
  author    = {Ye, Jiabo and Tian, Junfeng and Yan, Ming and others},
  booktitle = {Proc. IEEE/CVF Conf. Comput. Vis. Pattern Recognit. (CVPR)},
  pages     = {15502--15512},
  year      = {2022}
}

@inproceedings{HiVG,
  title     = {Hivg: Hierarchical multimodal fine-grained modulation for visual grounding},
  author    = {Xiao, Linhui and Yang, Xiaoshan and Peng, Fang and Wang, Yaowei and Xu, Changsheng},
  booktitle = {Proc. ACM Int. Conf. Multimedia (MM)},
  pages     = {5460--5469},
  year      = {2024}
}

@inproceedings{VG-LAW,
  title     = {Language adaptive weight generation for multi-task visual grounding},
  author    = {Su, Wei and Miao, Peihan and Dou, Huanzhang and others},
  booktitle = {Proc. IEEE/CVF Conf. Comput. Vis. Pattern Recognit. (CVPR)},
  pages     = {10857--10866},
  year      = {2023}
}

@inproceedings{SegVG,
  title        = {Segvg: Transferring object bounding box to segmentation for visual grounding},
  author       = {Kang, Weitai and Liu, Gaowen and Shah, Mubarak and Yan, Yan},
  booktitle    = {Proc. Eur. Conf. Comput. Vis. (ECCV)},
  pages        = {57--75},
  year         = {2024}
}

@article{LoRA,
  title   = {Lora: Low-rank adaptation of large language models.},
  author  = {Hu, Edward J and  others},
  journal = {in Proc. Int. Conf. Learn. Represent. (ICLR)},
  volume  = {1},
  number  = {2},
  pages   = {3},
  year    = {2022}
}

@inproceedings{CLIP-LoRA,
  title     = {Low-rank few-shot adaptation of vision-language models},
  author    = {Zanella, Maxime and Ben Ayed, Ismail},
  booktitle = {Proc. IEEE/CVF Conf. Comput. Vis. Pattern Recognit. (CVPR)},
  pages     = {1593--1603},
  year      = {2024}
}

@article{vilbert,
  title   = {Vilbert: Pretraining task-agnostic visiolinguistic representations for vision-and-language tasks},
  author  = {Lu, Jiasen and Batra, Dhruv and Parikh, Devi and Lee, Stefan},
  journal = {in Proc. Adv. Neural Inf. Process. Syst. (NeurIPS)},
  volume  = {32},
  year    = {2019}
}

@article{LLaMA2,
  title   = {Llama 2: Open foundation and fine-tuned chat models},
  author  = {Touvron, Hugo and Martin, Louis and Stone, Kevin and others},
  journal = {arXiv preprint arXiv:2307.09288},
  year    = {2023}
}

@inproceedings{ViLT,
  title        = {Vilt: Vision-and-language transformer without convolution or region supervision},
  author       = {Kim, Wonjae and Son, Bokyung and Kim, Ildoo},
  booktitle    = {Proc. Int. Conf. Mach. Learn. (ICML)},
  pages        = {5583--5594},
  year         = {2021},
  organization = {PMLR}
}

@article{TransVG++,
  title     = {Transvg++: End-to-end visual grounding with language conditioned vision transformer},
  author    = {Deng, Jiajun and Yang, Zhengyuan and Liu, Daqing and others},
  journal   = {IEEE Trans. Pattern Anal. Mach. Intell.},
  volume    = {45},
  number    = {11},
  pages     = {13636--13652},
  year      = {2023},
  publisher = {IEEE}
}

@article{ViT,
  title   = {An image is worth 16x16 words: Transformers for image recognition at scale},
  author  = {Dosovitskiy, Alexey and Beyer, Lucas and Kolesnikov, Alexander and others},
  journal = {arXiv preprint arXiv:2010.11929},
  year    = {2020}
}

@inproceedings{CLIP,
  title        = {Learning transferable visual models from natural language supervision},
  author       = {Radford, Alec and Kim, Jong Wook and Hallacy, Chris and others},
  booktitle    = {Proc. Int. Conf. Mach. Learn. (ICML)},
  pages        = {8748--8763},
  year         = {2021},
  organization = {PMLR}
}

@article{BERT,
  title   = {Bert: Pre-training of deep bidirectional transformers for language understanding},
  author  = {Devlin, Jacob and Chang, Ming-Wei and Lee, Kenton and Toutanova, Kristina},
  journal = {arXiv preprint arXiv:1810.04805},
  year    = {2018}
}

@inproceedings{UniTAB,
  title        = {Unitab: Unifying text and box outputs for grounded vision-language modeling},
  author       = {Yang, Zhengyuan and Gan, Zhe and Wang, Jianfeng and others},
  booktitle    = {Proc. Eur. Conf. Comput. Vis. (ECCV)},
  pages        = {521--539},
  year         = {2022}
}

@article{RVG-Tree,
  author     = {Hong, Richang and Liu, Daqing and Mo, Xiaoyu and He, Xiangnan and Zhang, Hanwang},
  title      = {Learning to Compose and Reason with Language Tree Structures for Visual Grounding},
  year       = {2022},
  issue_date = {Feb. 2022},
  publisher  = {IEEE Computer Society},
  address    = {USA},
  volume     = {44},
  number     = {2},
  issn       = {0162-8828},
  journal    = {IEEE Trans. Pattern Anal. Mach. Intell.},
  month      = feb,
  pages      = {684–696},
  numpages   = {13}
}

@article{CLIP-VG,
  author   = {Xiao, Linhui and Yang, Xiaoshan and Peng, Fang and others},
  journal  = {IEEE Trans. Multimedia},
  title    = {CLIP-VG: Self-Paced Curriculum Adapting of CLIP for Visual Grounding},
  year     = {2024},
  volume   = {26},
  number   = {},
  pages    = {4334-4347},
  doi      = {10.1109/TMM.2023.3321501}
}

@article{SimVG,
  title   = {Simvg: A simple framework for visual grounding with decoupled multi-modal fusion},
  author  = {Dai, Ming and Yang, Lingfeng and Xu, Yihao and Feng, Zhenhua and Yang, Wankou},
  journal = {in Proc. Adv. Neural Inf. Process. Syst. (NeurIPS)},
  volume  = {37},
  pages   = {121670--121698},
  year    = {2024}
}

@inproceedings{DQ-DETR,
  title        = {Dq-detr: Detr with dynamic query for tiny object detection},
  author       = {Huang, Yi-Xin and Liu, Hou-I and Shuai, Hong-Han and Cheng, Wen-Huang},
  booktitle    = {Proc. Eur. Conf. Comput. Vis. (ECCV)},
  pages        = {290--305},
  year         = {2024}
}

@inproceedings{Grounding-DINO,
  title        = {Grounding dino: Marrying dino with grounded pre-training for open-set object detection},
  author       = {Liu, Shilong and Zeng, Zhaoyang and Ren, Tianhe and others},
  booktitle    = {Proc. Eur. Conf. Comput. Vis. (ECCV)},
  pages        = {38--55},
  year         = {2024}
}

@article{groundinggpt,
  title   = {Groundinggpt: Language enhanced multi-modal grounding model},
  author  = {Li, Zhaowei and Xu, Qi and Zhang, Dong and others},
  journal = {arXiv preprint arXiv:2401.06071},
  year    = {2024}
}

@inproceedings{OFA,
  title        = {Ofa: Unifying architectures, tasks, and modalities through a simple sequence-to-sequence learning framework},
  author       = {Wang, Peng and others},
  booktitle    = {Proc. Int. Conf. Mach. Learn. (ICML)},
  pages        = {23318--23340},
  year         = {2022},
  organization = {PMLR}
}

@article{team2023gemini,
  title   = {Gemini: a family of highly capable multimodal models},
  author  = {Team, Gemini and Anil, Rohan and Borgeaud, Sebastian and others},
  journal = {arXiv preprint arXiv:2312.11805},
  year    = {2023}
}

@article{team2024gemini1.5,
  title   = {Gemini 1.5: Unlocking multimodal understanding across millions of tokens of context},
  author  = {Team, Gemini and Georgiev, Petko and Lei, Ving Ian and others},
  journal = {arXiv preprint arXiv:2403.05530},
  year    = {2024}
}

@inproceedings{autonomous_driving,
  title     = {Planning-oriented autonomous driving},
  author    = {Hu, Yihan and Yang, Jiazhi and Chen, Li and others},
  booktitle = {Proc. IEEE/CVF Conf. Comput. Vis. Pattern Recognit. (CVPR)},
  pages     = {17853--17862},
  year      = {2023}
}
